\documentclass[letterpaper]{article} 
\usepackage{aaai2026}  
\usepackage{times}  
\usepackage{helvet}  
\usepackage{courier}  
\usepackage[hyphens]{url}  
\usepackage{graphicx} 
\usepackage{natbib}  
\usepackage{caption} 
\usepackage{algorithm}
\usepackage{algpseudocode}
\usepackage{amsmath}
\usepackage{amssymb}
\usepackage{graphicx}
\usepackage{float}
\usepackage{multirow}
\usepackage{booktabs}
\usepackage{arydshln}
\usepackage{soul}

\usepackage[toc,page]{appendix}
\usepackage{hyperref}
\usepackage{booktabs}
\usepackage{array}
\usepackage{caption}
\usepackage[top=2cm, bottom=2cm, left=2cm, right=2cm]{geometry}
\usepackage[most]{tcolorbox}
\usepackage{xcolor}
\usepackage{fontawesome}

\usepackage{newfloat}
\usepackage{listings}

\tcbuselibrary{breakable}
\tcbuselibrary{raster}
\tcbuselibrary{skins}

\tcbset{
    enforce breakable
}

\newcommand{\wrong}{\textcolor{red}{\faTimes}}
\newcommand{\correct}{\textcolor{green!70!black}{\faCheck}}

\addcontentsline{toc}{section}{Appendices}

\DeclareCaptionStyle{ruled}{labelfont=normalfont,labelsep=colon,strut=off} 
\floatstyle{ruled}
\newfloat{listing}{tb}{lst}{}
\floatname{listing}{Listing}
\title{Tool-Augmented Policy Optimization: Synergizing Reasoning and Adaptive Tool Use with Reinforcement Learning}
\author{
    Wenxun Wu\textsuperscript{\rm 1, 2},
    Yuanyang Li\textsuperscript{\rm 1},
    Guhan Chen\textsuperscript{\rm 3},
    Linyue Wang\textsuperscript{\rm 1},
    Hongyang Chen\textsuperscript{\rm 2}
}
\affiliations{
    \textsuperscript{\rm 1}Zhejiang University\\
    \textsuperscript{\rm 2}Zhejiang Laboratory \\
    \textsuperscript{\rm 3}Tsinghua University \\

}

\usepackage{bibentry}

\begin{document}

\maketitle

\begin{abstract}
Recent advances in large language models (LLMs) have popularized test-time scaling, where models generate additional reasoning tokens before producing final answers. These approaches have demonstrated significant performance improvements on benchmarks involving mathematical reasoning. However, language models relying solely on direct inference still struggle with tasks demanding \textbf{up-to-date knowledge} or \textbf{computational tools} such as calculators and code interpreters for complex arithmetic operations. To overcome these limitations, we propose \textbf{T}ool-\textbf{A}ugmented \textbf{P}olicy \textbf{O}ptimization (\textbf{TAPO}), a novel reinforcement learning framework that systematically integrates multi-hop reasoning with adaptive tool-calling capabilities. Our approach employs a modified version of Dynamic Sampling Policy Optimization (DAPO), a recently developed RL paradigm, which we adapt specifically for tool invocation scenarios, enabling models to dynamically interleave complex reasoning with on-demand tool usage (including search APIs and Python interpreters).

To support this research, we introduce two new datasets: \texttt{TAPO-easy-60K} and \texttt{TAPO-hard-18K}, specifically designed to train and evaluate both fact-based reasoning and mathematical calculation capabilities. Our experiments on Qwen2.5-3B and Qwen2.5-7B models demonstrate the effectiveness of our approach, with both models achieving state-of-the-art performance on tasks requiring external knowledge and mathematical computation among methods with comparable parameters. Notably, TAPO achieves more efficient tool utilization than baseline methods while preventing excessive calls caused by reward hacking. These results highlight the significant potential of combining advanced reasoning with tool usage to enhance model performance in knowledge-intensive and computationally demanding tasks.

Our code is available at \href{https://github.com/Goer17/TAPO}{https://github.com/Goer17/TAPO}.
\end{abstract}

\section{Introduction}
The field of large language models (LLMs) has seen significant advancements across diverse capabilities, from fundamental language understanding to complex reasoning and generation tasks. A key development in this evolution has been the emergence of intermediate reasoning tokens—where models generate step-by-step rationales before final answers—which has proven particularly effective for enhancing performance \cite{wei2022chain,lightman2023let}. Subsequent research has formalized this approach through test-time scaling laws, demonstrating that systematically increasing reasoning tokens yields substantial improvements on challenging benchmarks \cite{snellScalingLLMTestTime2024}.

Reinforcement learning has emerged as a powerful paradigm for optimizing these reasoning capabilities. Recent commercial LLMs such as OpenAI-o1 \cite{jaech2024openai} and DeepSeek-R1 \cite{guo2025deepseek} have successfully leveraged RL algorithms—including Proximal Policy Optimization (PPO) \cite{schulman2017proximal} or Group Relative Policy Optimization (GRPO) \cite{shao2024deepseekmath}—to maximize the benefits of test-time scaling while maintaining output quality.

Despite significant advances in reasoning capabilities, large language models (LLMs) still face fundamental limitations when handling tasks that require up-to-date knowledge or complex numerical computation. While traditional approaches like Retrieval-Augmented Generation (RAG)~\cite{lewis2020retrieval} address knowledge gaps through external retrieval, and function calling enables tool interaction, these methods suffer from critical shortcomings: (1) they lack explicit intermediate reasoning steps, and (2) they struggle with multi-hop tool invocation---where iterative reasoning should guide dynamic tool selection.

These limitations have spurred recent efforts to combine LLM reasoning with tool augmentation via RL. Notable approaches include \textsc{Search-R1}~\cite{jin2025search}, which interleaves reasoning with search engine queries for improved retrieval, and \textsc{ReTool}~\cite{fengReToolReinforcementLearning2025}, which augments models with code interpreters. Although these methods demonstrate strong in-domain performance, we identify two key shortcomings: (1) \textbf{significant generalization gaps across tasks}—for example, a search-engine-augmented language model may exhibit a sharp decline in mathematical reasoning ability, performing even worse than its baseline; and (2) \textbf{reward hacking} behaviors, particularly excessive tool calls, when trained on narrow task categories.

In this work, we introduce \textbf{Tool-Augmented Policy Optimization (TAPO)}, an enhanced training framework derived from Dynamic Sampling Policy Optimization (DAPO)~\cite{yu2025dapo}. Our TAPO-trained LLM integrates:
\begin{itemize}
    \item Multi-step reasoning
    \item Enhanced tool invocation capabilities:
    \begin{itemize}
        \item \textbf{Search engine} for real-time knowledge retrieval
        \item \textbf{Remote Python interpreter} for computational tasks
    \end{itemize}
\end{itemize}
This unified framework enables seamless coordination between reasoning processes and tool utilization.

Following the prompt design of DeepSeek-R1, our system employs structured XML-like formatting with key tags: \texttt{<think>} for reasoning, \texttt{<search>} for queries, \texttt{<code>} for Python execution,\texttt{<response>} for external outputs, and \texttt{<answer>} for final conclusions.

To train our model, we prepared two high-quality datasets: \texttt{TAPO-easy-60K} (about 60,000 samples) and \texttt{TAPO-hard-18K} (about 18,000 samples), collectively containing 78K question-answer pairs spanning both fact-reasoning and mathematics problems - with both dataset types containing examples of each problem category. We designed specialized reward functions tailored to different question types to ensure proper training signals.

In summary, our contributions include:
\begin{enumerate}
\item \textbf{TAPO, a novel RL framework} for training language models with interleaved reasoning and tool invocation (supporting search engines, code interpreters, and extensible to other APIs);
\item \textbf{Two high-quality datasets} (\texttt{TAPO-easy-60K} and \texttt{TAPO-hard-18K}) for training and evaluating knowledge retrieval and mathematical computation;
\item Optimized 7B/3B models achieving \textbf{state-of-the-art comprehensive abilities} with \textbf{high tool invocation efficiency}.
\end{enumerate}

\section{Related Work}
\subsection{Test-Time Scaling Law in LLMs}

Traditionally, researchers primarily enhanced large language model (LLM) performance through scaling model size, dataset size, and training compute budgets~\cite{kaplanScalingLawsNeural2020}. However, recent studies have demonstrated that scaling test-time computation—allowing LLMs to generate more intermediate tokens before producing final outputs—can yield superior efficiency gains~\cite{snellScalingLLMTestTime2024,wuINFERENCESCALINGLAWS2025}. 

This phenomenon parallels human cognitive processes: just as humans expend more time deliberating on complex problems, LLMs benefit from extended reasoning chains. The incorporation of chain-of-thought (CoT)~\cite{wei2022chain} as an intermediate reasoning stage has proven particularly effective for improving answer quality on challenging tasks, as it enables more thorough information processing before final response generation.

\subsection{Planning, Reasoning and Acting in LLMs}

While test-time scaling improves reasoning through extended token generation~\cite{snellScalingLLMTestTime2024}, Transformer-based LLMs~\cite{vaswani2017attention} still face: (1) hallucination in precise tasks, and (2) static knowledge requiring retraining. Hybrid approaches like ReAct~\cite{yaoReActSynergizingReasoning2023} (reasoning+tools) and LATS~\cite{zhou2023language} (with MCTS) partially address these through external actions such as RAG~\cite{lewis2020retrieval} or code execution, but remain constrained by supervised paradigms.

\subsection{Reinforcement Learning in LLMs}

In recent years, reinforcement learning (RL) has emerged as a powerful paradigm for enhancing large language model (LLM) capabilities, with both off-policy and on-policy approaches demonstrating significant potential. While off-policy methods like Direct Preference Optimization (DPO)~\cite{rafailov2023direct}, Simple Preference Optimization (SimPo)~\cite{meng2024simpo} and Kahneman-Tversky Optimization (KTO)~\cite{ethayarajhKTOModelAlignment2024} can achieve strong performance through preference-based learning, they often require large-scale high-quality preference datasets that are challenging to acquire. In contrast, on-policy methods such as Proximal Policy Optimization (PPO)~\cite{schulman2017proximal} and Group Relative Policy Optimization (GRPO)~\cite{shao2024deepseekmath} offer more practical training pipelines by generating their own training data through policy-environment interaction, albeit with increased computational demands. These on-policy techniques have proven particularly effective for improving LLM performance on complex reasoning tasks.










\subsubsection{Group Relative Policy Optimization (GRPO)}

Group Relative Policy Optimization (GRPO) \cite{shao2024deepseekmath} is a lightweight reinforcement learning framework that employs group-based trajectory sampling and intra-group advantage computation. The method first samples a batch of trajectories from the current policy, then calculates relative advantage functions exclusively within each sampled group. The optimization objective is formally expressed as:

\begin{equation}
\begin{aligned}
&\mathcal{J}_\text{GRPO}(\theta) = \mathbb{E}_{(q, a) \sim D , o \sim \pi_{\theta_\text{old}}} \Biggl[ \frac{1}{G} \sum_{i=1}^G\frac{1}{|o_i|}\sum_{t=1}^{|o_i|} \\
&\quad \Bigl(\min (r_{i,t}(\theta)A_i, \text{clip}(r_{i,t}(\theta),1-\epsilon, 1+\epsilon)A_i\Bigr) \\
&\quad - \beta D_{KL}(\pi_\theta || \pi_\text{ref}) \Biggr]
\end{aligned}
\end{equation}

Where $r_{i,t}(\theta) = \frac{\pi_\theta(o_{i,t}|q,o_{i,<t})}{\pi_{\theta_\text{old}}(o_{i,t}|q,o_{i,<t})}$ is the policy probability ratio, and the advantage $A_i$ is computed through z-score normalization within each group of samples:
\begin{equation}
    A_i = \frac{R_i - \text{mean}(\{R_j\}_{j=1}^G)}{\text{std}(\{R_j\}_{j=1}^G)}
\end{equation}

\subsubsection{Dynamic Sampling Policy Optimization (DAPO)}
 Dynamic Sampling Policy Optimization (DAPO) \cite{yu2025dapo} is a novel reinforcement learning paradigm that addresses two critical limitations of GRPO \cite{shao2024deepseekmath}: (1) the policy model training inefficiency caused by degenerate advantage computation scenarios where z-score normalization becomes meaningless when a sample group contains uniformly correct or incorrect responses, and (2) the significant entropy collapse observed during GRPO training processes.

To overcome GRPO's limitations, DAPO introduces five key innovations: (1) \textbf{Dynamic sampling} as the algorithmic core, which ensures diverse sample quality within each batch by strategically mixing high- and low-quality samples, guaranteeing valid advantage computation and significantly accelerating training compared to GRPO; (2) \textbf{Asymmetric clipping} with decoupled bounds ($\epsilon_\text{low}$, $\epsilon_\text{high}$) for advantage function stabilization, where specifically $\epsilon_\text{high} > \epsilon_\text{low}$ prevents entropy collapse while maintaining training stability; and (3) \textbf{Adaptive length penalty}, a soft reward shaping mechanism that combines the base correctness reward with a length-aware penalty term, effectively discouraging verbose outputs while preserving response quality. (4) \textbf{KL Penalty Elimination}: DAPO completely removes the $\beta D_{KL}(\pi_\theta \parallel \pi_{\text{ref}})$ term from the loss calculation, simplifying the optimization objective while maintaining stability through its dynamic sampling mechanism; (5) \textbf{Token-Level Gradient Computation}: Unlike GRPO's sample-level policy updates, DAPO implements fine-grained $\nabla_\theta \mathcal{J}(\theta)$ computation at each token position $t$ in the sequence $o_i$. The objective function of DAPO is formulated as:

\begin{equation}
\begin{aligned}
&\mathcal{J}_\text{DAPO}(\theta) = \mathbb{E}_{(q, a) \sim D , o \sim \pi_{\theta_\text{old}}} \\
&\quad \Biggl[ \frac{1}{\sum_{i=1}^G|o_i|} \sum_{i=1}^G \sum_{t=1}^{|o_i|} \\
&\quad \Bigr(\min(r_{i,t}(\theta)A_i, \text{clip}(r_{i,t}(\theta),1-\epsilon_\text{low}, 1+\epsilon_\text{high})A_i\Bigr) \Biggr]
\end{aligned}
\end{equation}

Where:

\begin{equation}
    \text{std}\Bigl(\{R_i\}_{i=1}^G\Bigr) > 0
\end{equation}

\section{Our Work: Tool-Augmented Policy Optimization}
We propose \textbf{T}ool-\textbf{A}ugmented \textbf{P}olicy \textbf{O}ptimization (\textbf{TAPO}), a novel reinforcement learning framework that systematically coordinates language models' \textbf{reasoning} and \textbf{tool manipulation} capabilities through \textbf{on-policy reinforcement learning}. This section introduces: (1) an enhanced inference mechanism that dynamically interleaves token generation with tool invocation, and (2) an RL training paradigm that jointly optimizes the policy's reasoning quality and tool utilization.

\subsection{Inference with Multi-turn Tool Calling}

Prior research has demonstrated two complementary approaches to augmenting LLM capabilities: search engine integration for knowledge retrieval \cite{nakano2021webgpt,wu2025webdancer,jin2025search} and inline code execution for complex computations \cite{schick2023toolformer,fengReToolReinforcementLearning2025}, which respectively address factual reasoning and numerical processing limitations.

\begin{figure*}[t]
    \includegraphics[width=\textwidth]{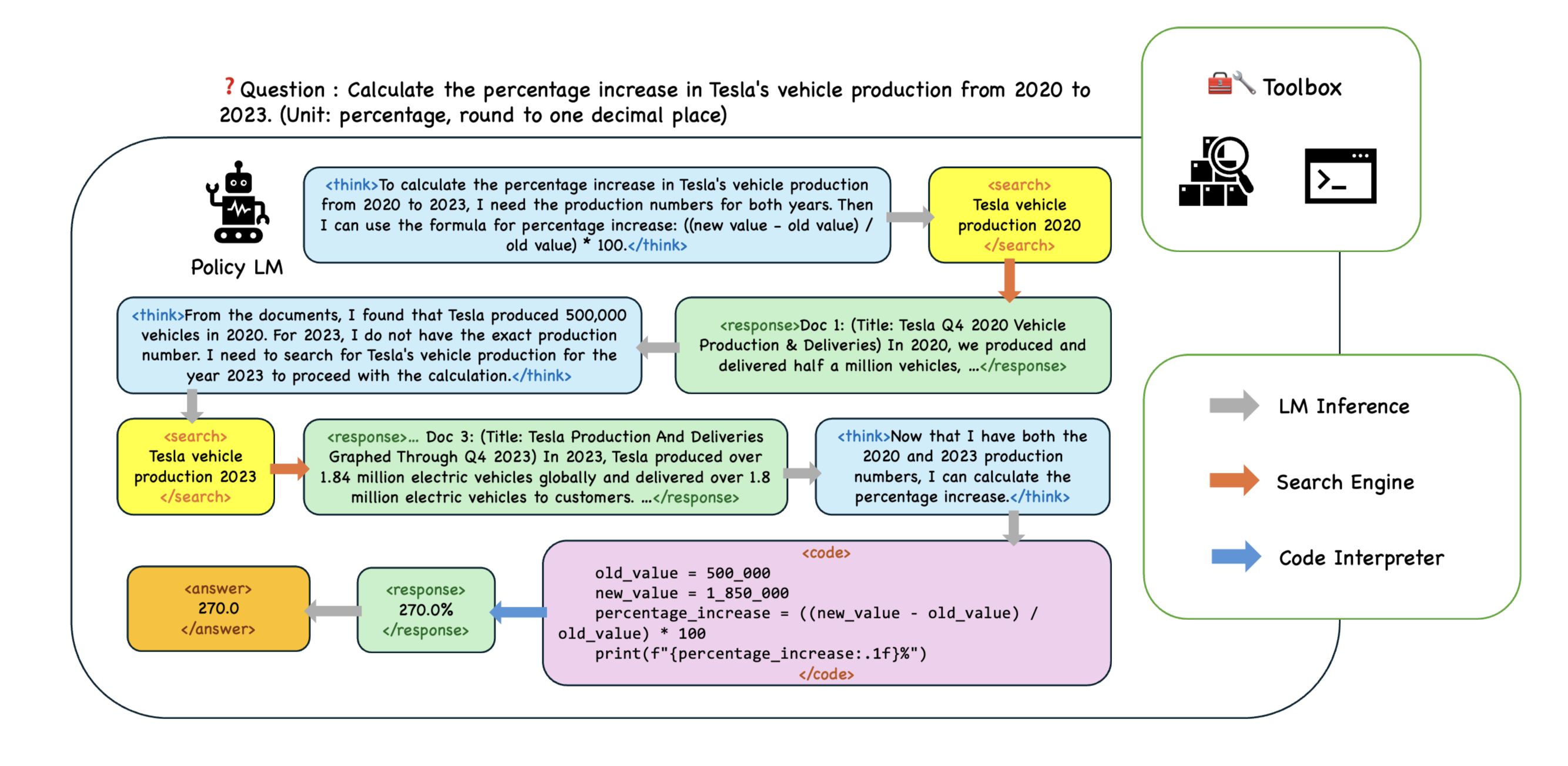}
    \caption{An Example of Policy Model Inference in TAPO. The policy language model generates multi-stage reasoning outputs while automatically appending tool-calling responses (e.g., from search engine or Python code interpreter) when external tools are invoked.}
    \label{fig:inference}
\end{figure*}

Our work advances beyond these foundations by simultaneously integrating both search APIs and Python interpreters within a unified framework, while introducing (1) interleaved reasoning traces generated prior to tool invocation, and (2) an XML-structured output format adopting DeepSeek-R1's paradigm \cite{guo2025deepseek} with specialized markup tags: reasoning steps encapsulated in \texttt{<think>}...\texttt{</think>} pairs, tool calls delimited by \texttt{<tool-name>}...\texttt{</tool-name>} tags, and tool responses wrapped in \texttt{<response>}...\texttt{</response>} sections, thereby establishing explicit phase boundaries between reasoning, tool invocation, and result integration (shown in Algorithm \ref{alg:infer} and Figure \ref{fig:inference}).

\begin{algorithm}[tb]
\caption{Tool-Augmented Inference}
\label{alg:infer}
\textbf{Input}: prefix prompt $q$ including system prompt and question, policy model $\theta$, search engine SE, code interpreter CI and the maximum number of tool calling max\_turns. \\
\textbf{Output}: inference trajectory $O$.

\begin{algorithmic}[1]
\State $O\gets \texttt{""}$, $\text{response} \gets \texttt{""}$
\For{$i \gets 1$ \textbf{to} $\text{max\_turns} + 1$}
    \State $t \gets O.\text{length}$
    \While{$t < \text{max\_response\_length}$}
        \State $o_t \gets \text{LLM}_\theta(o_t \mid q, o_{<t})$
        \State $O \gets O + o_t$
        \State $t \gets t + 1$
        \If{$O.\text{endswith}(\texttt{</answer>})$}
            \State \textbf{return} $O$
        \EndIf
        \If{$O.\text{endswith}(\texttt{</search>})$}
            \State $Q \gets O.\text{match}(\texttt{<search>*</search>})[-1]$
            \State response $\gets$ SE($Q$)
            \State \textbf{break}
        \ElsIf{$O.\text{endswith}(\texttt{</code>})$}
            \State $C \gets O.\text{match}(\texttt{<code>*</code>})[-1]$
            \State response $\gets$ CI($C$)
            \State \textbf{break}
        \EndIf
    \EndWhile
    \State $O \gets O$ + \texttt{<response>}response\texttt{</response>}
\EndFor
\State \textbf{return} $O$
\end{algorithmic}
\end{algorithm}

\subsubsection{Search Engine}
Our search system combines the Google Serper API \cite{serper2023} with Redis caching to balance real-time retrieval and efficiency. The two-tiered strategy first attempts exact cache matches, then activates fuzzy matching via Python's \texttt{difflib.SequenceMatcher} (threshold $>0.9$) on normalized (lowercase, whitespace-removed) queries. This reduces API calls while delivering millisecond responses - crucial for reinforcement learning's iterative training demands.

\subsubsection{Code Interpreter}
We implement a high-performance Python Code Interpreter (CI) as a remote HTTP service using FastAPI, designed for secure and scalable code execution. The server architecture features:
\begin{itemize}
    \item Multi-core parallel processing capable of handling 10,000 requests per second. (verified through throughput testing)
    \item Docker containerization for environment isolation and security hardening.
    \item Strict package whitelisting (e.g., \texttt{math}, \texttt{sympy}, \texttt{itertools}) with blacklisted system modules (e.g., \texttt{os}, \texttt{sys}, file I/O operations).
    \item Execution time limit of 5,000ms (5 seconds) with automatic timeout enforcement.
\end{itemize}
This sandboxed implementation prevents infinite loops, restricts unsafe operations, and maintains system stability during concurrent code execution.

\subsection{Reinforcement Learning with Tool Calling}

The TAPO framework implements an on-policy reinforcement learning algorithm derived from the DAPO paradigm~\cite{yu2025dapo}. Our training process consists of two alternating phases (shown in figure \ref{fig:tapo}):

\begin{figure*}[t]
    \centering
    \includegraphics[width=1\linewidth]{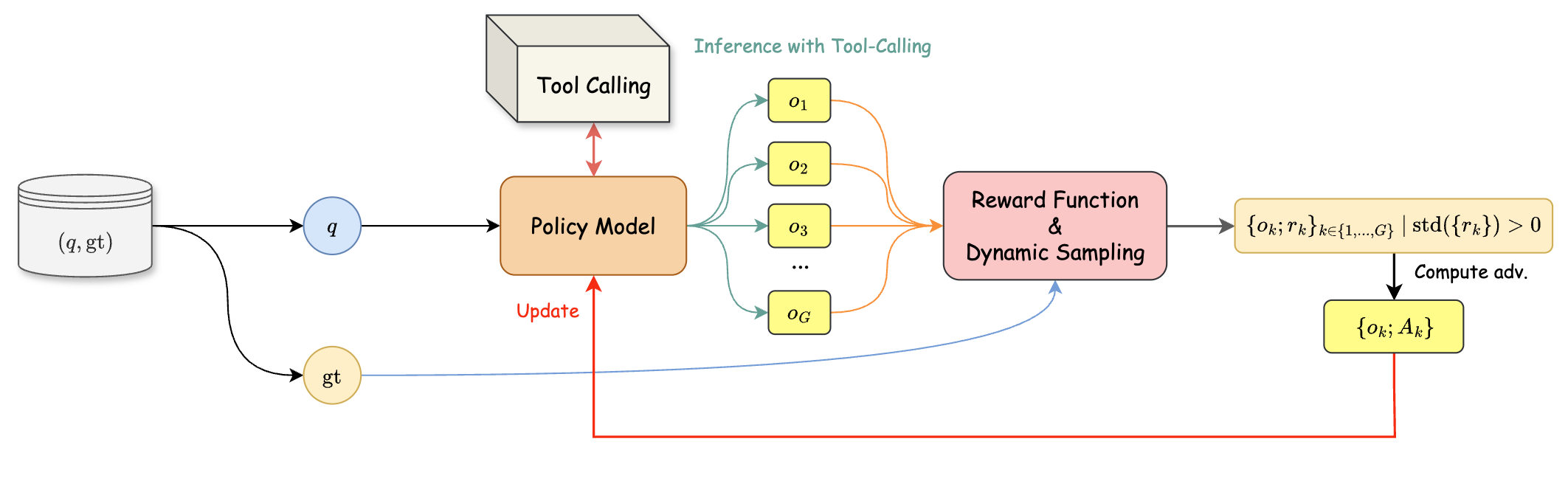}
    \caption{Overview of the TAPO pipeline. (1) The policy model first generates a batch of tool-augmented responses, which are then scored based on ground truths. (2) Responses are dynamically sampled from groups with non-zero standard deviation to ensure diversity. (3) Finally, the advantage function is computed and used to update the policy model.}
    \label{fig:tapo}
\end{figure*}

\begin{itemize}
    \item \textbf{Rollout with Tool Calling:}
    The policy model generates a batch of responses by executing tool-augmented reasoning chains, following the inference rules described in previous section.
    \item \textbf{Policy Optimization:}
    Upon finishing a rollout phase, the advantage function is calculated using the trajectories gathered through dynamic sampling~\cite{yu2025dapo}. The policy parameters $\theta$ are then updated through gradient ascent on the target objective:
\end{itemize}

\begin{equation}
\begin{aligned}
& \mathcal{J}_\text{TAPO}(\theta) = \mathbb{E}_{(q, a) \sim D, o \sim (\pi_{\theta_\text{old}}, \text{tool})} \\
& \Biggl[\frac{1}{\sum_{i = 1}^G\sum_{t = 1}^{|o_i|} \mathbb{I}(o_{i,t})} \sum_{i = 1}^G \sum_{t = 1}^{|o_i|} \mathbb{I}(o_{i,t}) \\
& \min\Bigl(r_{i,t}(\theta) A_i, \text{clip}(r_{i, t}(\theta), 1 - \epsilon_\text{low}, 1 + \epsilon_\text{high}) A_i\Bigr)\Biggr]
\end{aligned}
\end{equation}

Where $r_{i,t}(\theta) = \pi_\theta(o_{i,t}|q,o_{i,<t})/\pi_{\theta_\text{old}}(o_{i,t}|q,o_{i,<t})$ is the policy probability ratio, and $A_i = (R_i - \mu_G)/\sigma_G$ is the standardized advantage computed over dynamic groups ($\mu_G,\sigma_G$ denote group mean/std of returns $\{R_j\}_{j=1}^G$).

\subsubsection{Response Masking for Tool Calling}
To isolate the language model's contribution during training, we introduce a binary mask $\mathbb{I}(o_{i,t})$ that excludes tool-generated tokens from the loss computation. The mask is defined as:

\begin{equation}
    \mathbb{I}(o_{i,t}) = \begin{cases}
    1 & \text{if token } o_{i,t} \text{ is generated by language model} \\
    0 & \text{if token } o_{i,t} \text{ is produced by tool execution}
    \end{cases}
\end{equation}

This masking ensures the optimization function $\mathcal{J}(\theta)$ only updates the model parameters based on its own generated tokens:

\subsubsection{Rule-based Rewarding}

TAPO's reward function combines three critical aspects of response quality: format compliance, answer accuracy, and length efficiency. The accuracy component first verifies output formatting and correctness:

\begin{equation}
    R_\text{acc}(o, gt, \text{T}) = \begin{cases}
        -1 & \text{if } o \text{ has incorrect format} \\
        f_\text{T}(o, gt) & \text{otherwise}
    \end{cases}
\end{equation}

Here, $o$ denotes the model's output, $gt$ the ground truth, and $\text{T}$ the question type. The type-specific scorer $f_\text{T}: (o, gt) \to [0, 1]$ adapts to different evaluation needs. Unlike rigid approaches like Search-R1's exact matching \cite{jin2025search}—which would penalize valid variants like "Wilhelm Röntgen" versus the full "Wilhelm Conrad Röntgen"—our method preserves semantic correctness through specialized scoring functions.

For factual questions, we use a normalized edit-distance metric:
\begin{equation}
    f_\text{fact}(o, gt) = \begin{cases}
        0 & \text{if } \text{r}(o, gt) > 0.5 \\
        1 - \text{r}(o, gt) & \text{if } \text{r}(o, gt) \le 0.5
    \end{cases}
\end{equation}
where $\text{r}(o, gt) = \frac{\text{lev}(o, gt)}{\max(|o|, |gt|)}$, $\text{lev}(o, gt)$ is the string Levenshtein distance \cite{levenshtein1966binary} between $o$ and $gt$. Mathematical answers instead employ strict matching ($f_\text{math}(o, gt) = \mathbb{I}[o \equiv gt]$) due to their objective nature.

The length penalty, adapted from DAPO \cite{yu2025dapo}, focuses solely on policy-generated tokens:
\begin{equation}
    R_\text{length}(o) = \begin{cases}
        0 &  L_o \le L_\text{max} - L_\text{cache} \\
        \frac{L_\text{max} - L_\text{cache} - L_o}{L_\text{cache}} & L_\text{max} - L_\text{cache} < L_o < L_\text{max} \\
        -1 & L_o \ge L_\text{max}
    \end{cases}
\end{equation}

Where $L_o = \sum_{t=1}^{|o|}\mathbb{I}(o_t)$.

These components combine additively to form the final reward:
\begin{equation}
    \mathbb{R}(o, gt, \text{T}) = R_\text{acc}(o, gt, \text{T}) + R_\text{length}(o)
\end{equation}

\section{Experiments and Results}
In this section, we present the experimental framework and results of TAPO.

\subsection{Datasets: TAPO-easy-60K and TAPO-hard-18K}

To train our model, we introduce two high-quality question-answering datasets encompassing both knowledge-intensive and mathematical computation tasks: \textbf{TAPO-easy-60K} and \textbf{TAPO-hard-18K}.

\subsubsection{TAPO-easy-60K}

\texttt{TAPO-easy-60K} contains approximately 27,000 mathematical problems and 33,000 fact-reasoning questions. The mathematical problems are primarily sourced from GSM8K \cite{hendrycks2021measuring} and DAPO-MATH-17K \cite{yu2025dapo}. However, we observe that these existing datasets predominantly evaluate mathematical reasoning and logical capabilities, with less emphasis on precise numerical computation. To address real-world scenarios where models encounter problems requiring calculator usage—similar to computational challenges faced in practical applications—we developed \texttt{calculator-6K}, an automatically generated dataset of computation-intensive problems created using our Python scripts. This addition enhances the model's capability to handle tasks demanding accurate numerical calculations.

For fact-reasoning questions, we performed extensive curation of the NQ dataset \cite{kwiatkowski2019natural}, implementing rigorous filtering to ensure data quality. Specifically, we removed temporally ambiguous questions that lack explicit time constraints, such as queries asking "What is the GDP of a certain region this year?" without specifying the year. Such questions produce inconsistent answers across different time periods, potentially introducing negative training signals for search engine invocation. By eliminating these temporally sensitive questions, we ensure that our training data provides consistent and reliable supervision for tool usage learning.

\subsubsection{TAPO-hard-18K}
\texttt{TAPO-hard-18K} consists of 10,000 challenging questions carefully selected from DeepMath-103K, combined with our newly constructed \texttt{complex-8K} dataset containing 8,000 comprehensive questions requiring multi-hop retrieval and computation. To ensure high quality, we initially collected 20,000 candidate questions from various online sources and manually created additional samples. These questions were then evaluated using three commercial LLMs (OpenAI-GPT-4o \cite{hendrycks2024gpt4o}, DeepSeek-V3 \cite{liu2024deepseek}, and Gemini 2.5 Pro \cite{deepmind2025gemini}) equipped with our search API and code interpreter. Only questions correctly answered by all three models were retained, yielding approximately 8,000 high-quality QA pairs after human verification. These final questions demand both external knowledge retrieval and complex computational capabilities provided by Python code interpretation.

\subsection{Experimental Setup}
Our experiments employ two model variants: \textsc{Qwen2.5-7B} and \textsc{Qwen2.5-3B}~\cite{qwen2024}. Both models are trained on a unified dataset combining \texttt{TAPO-easy-60K} and \texttt{TAPO-hard-18K} (78K samples total), with 66K samples for training and the remaining 12K for testing. The 7B variant uses batch size 128, sample group size 8, and maximum response length 8192 tokens, while the 3B variant employs batch size 64 with the same group size but shorter maximum length (4096 tokens). Both configurations share identical tool-calling budgets (4 calls per rollout) and clip parameters ($\epsilon_{\text{low}} = 0.2$, $\epsilon_{\text{high}} = 0.28$), following \textsc{DAPO}~\cite{yu2025dapo}.

\subsection{Performance}

We trained the 7B and 3B models separately (2 epochs each) on 8 NVIDIA A100 80GB GPUs, enabling both search engine and code interpreter tool capabilities. Both models demonstrated remarkable performance compared to other models of similar scale, achieving competitive results across all evaluation metrics (detailed comparisons are provided in Table \ref{tab:performance}).

To further evaluate the generalization capability of our method, we test our TAPO-trained 7B model on three out-of-domain (OOD) benchmarks: HotPotQA~\citep{yang2018hotpotqadatasetdiverseexplainable}, TriviaQA~\citep{joshi2017triviaqalargescaledistantly}, and MATH~\cite{hendrycks2021measuringmathematicalproblemsolving}. As shown in Table~\ref{tab:ood-performance}, our approach achieves competitive generalization performance with a 7B parameter model.

\begin{table*}[t]
\centering
\small
\setlength{\tabcolsep}{4pt}
\begin{tabular}{@{}lcccccccl@{}}
\toprule
\multirow{2}{*}{Methods} & \multicolumn{1}{c}{Fact-reasoning} & \multicolumn{4}{c}{Math} & Comprehensive & \multicolumn{2}{c}{Avg.} \\
\cmidrule(lr){2-2} \cmidrule(lr){3-6} \cmidrule(lr){7-7} \cmidrule(lr){8-9}
 & NQ$^\dagger$ & Calculator-6K$^\dagger$ & GSM8K$^\dagger$ & DAPO-Math$^\dagger$ & DeepMath$^*$ & Complex-8K$^*$ & easy & hard \\
\midrule
Qwen2.5-7B-Instruct & 24.5 & 36.9 & 62.5 & 19.6 & 36.2 & 9.6 & 28.4 & 24.3 \\
Qwen2.5-7B-Instruct + SE/CI & 48.0 & 64.2 & 58.3 & 29.2 & 39.8 & 41.4 & 46.9 & 40.5 \\
Qwen2.5-Math-7B-Instruct & 4.2 & 44.0 & \textbf{95.4} & 30.5 & 46.0 & 1.2 & 26.0 & 26.1 \\
Qwen2.5-Math-7B-Instruct + SE/CI & 24.1 & 76.2 & 95.1 & 44.5 & 47.1 & 31.2 & 43.2 & 40.1 \\
Search-R1-7B & 51.0 & 22.3 & 26.9 & 16.3 & 11.1 & 13.9 & 37.4 & 12.3 \\
\textbf{TAPO-trained Qwen2.5-7B (ours)} & \textbf{52.1} & \textbf{98.6} & 92.2 & \textbf{47.1} & \textbf{54.1} & \textbf{68.7} & \textbf{61.1} & \textbf{60.6} \\
\midrule
Qwen-2.5-3B-Instruct & 12.9 & 16.4 & 41.4 & 13.8 & 21.9 & 9.1 & 17.2 & 16.2 \\
Qwen-2.5-3B-Instruct + SE/CI & 42.9 & 48.7 & 42.1 & 17.5 & 34.3 & 37.9 & 37.9 & 35.9 \\
Search-R1-3B & 40.3 & 14.9 & 12.6 & 9.8 & 24.3 & 26.8 & 27.5 & 25.4 \\
\textbf{TAPO-trained Qwen2.5-3B (ours)} & \textbf{46.9} & \textbf{98.0} & \textbf{85.6} & \textbf{38.5} & \textbf{39.9} & \textbf{46.4} & \textbf{55.4} & \textbf{42.8} \\
\bottomrule
\end{tabular}
\caption{Pass@1 accuracy (\%) comparison across different methods and datasets. SE and CI denote search engine and code interpreter tool augmentation respectively. Notably, Search-R1 is evaluated using our standardized search engine for fair comparison. The $^\dagger$ symbol indicates datasets from \texttt{TAPO-easy-60K}, while $^*$ denotes datasets from \texttt{TAPO-hard-18K}.}
\label{tab:performance}
\end{table*}

\begin{table*}[t]
\centering
\small
\setlength{\tabcolsep}{4pt}
\begin{tabular}{@{}lcccccccl@{}}
\toprule
\multirow{2}{*}{Methods} & \multicolumn{2}{c}{Fact-reasoning} & \multicolumn{1}{c}{Math} & \multicolumn{1}{c}{Avg.} \\
\cmidrule(lr){2-3} \cmidrule(lr){4-4} \cmidrule(lr){5-5} \\
 & HotpotQA & TriviaQA & MATH &  \\
\midrule
Qwen2.5-7B-Instruct + SE/CI & 39.8 & 61.1 & 62.8 & 54.5 \\
Qwen2.5-Math-7B-Instruct + SE/CI& 21.3 & 32.9 & \textbf{85.6} & 46.6 \\
Search-R1-7B & \textbf{44.2} & 62.7 & 25.7 & 41.2 \\
\textbf{TAPO-trained Qwen2.5-7B (our method)} & 40.1 & \textbf{65.7} & 81.2 & \textbf{62.3} \\
\bottomrule
\end{tabular}
\caption{Pass@1 accuracy (\%) of different methods across various out-of-domain datasets.}
\label{tab:ood-performance}
\end{table*}

\subsection{Ablation Study}

To systematically evaluate the individual contributions of each core component within our proposed framework, we conduct an extensive ablation study utilizing the \textsc{Qwen}2.5-3B model as the base architecture. This analysis is designed to isolate and quantify the performance impact of the reasoning module, the search engine tool, and the code interpreter by comparing the performance of the complete model against several strategically ablated variants. The corresponding results, which clearly demonstrate the necessity of each integrated component, are meticulously detailed in Table~\ref{tab:ablation}.

\begin{table}[H]
\centering
\small
\caption{Performance analysis of ablated \textsc{Qwen}2.5-3B variants (pass@1 accuracy). ↓ indicates performance drop, ↑ indicates gain relative to full model.}
\label{tab:ablation}
\begin{tabular}{lccc}
\toprule
\textbf{Variant} & \textbf{Fact-reasoning} & \textbf{Math} & \textbf{Comp} \\
\midrule
Full model & 46.9 & 60.5 & 46.4 \\
w/o reasoning & 45.7 ↓ & 52.3 ↓ & 39.9 ↓ \\
w/o search engine & 14.4 ↓ & 62.7 ↑ & 13.8 ↓ \\
w/o code interpreter & 46.1 ↓ & 42.8 ↓ & 36.0 ↓ \\
\bottomrule
\end{tabular}
\vspace{0.2cm}
\end{table}

The ablation results reveal some key findings: (1) The reasoning module provides crucial support for mathematical tasks, with its removal causing an 8.2\% performance drop; (2) The search engine proves indispensable for fact-reasoning, where its absence leads to a dramatic 32.5\% accuracy decrease; and (3) The code interpreter plays a vital role in mathematical reasoning, contributing to 17.7\% of the model's math performance. Additionally, both the search engine and code interpreter significantly impact comprehensive task performance. These components demonstrate complementary effects, with the mathematical reasoning capability providing a small boost to fact-reasoning performance, suggesting some cross-domain knowledge transfer within the policy model.

\subsection{Tool Invocation Frequency Analysis}

\begin{figure}[H]
\centering
\includegraphics[width=0.95\linewidth]{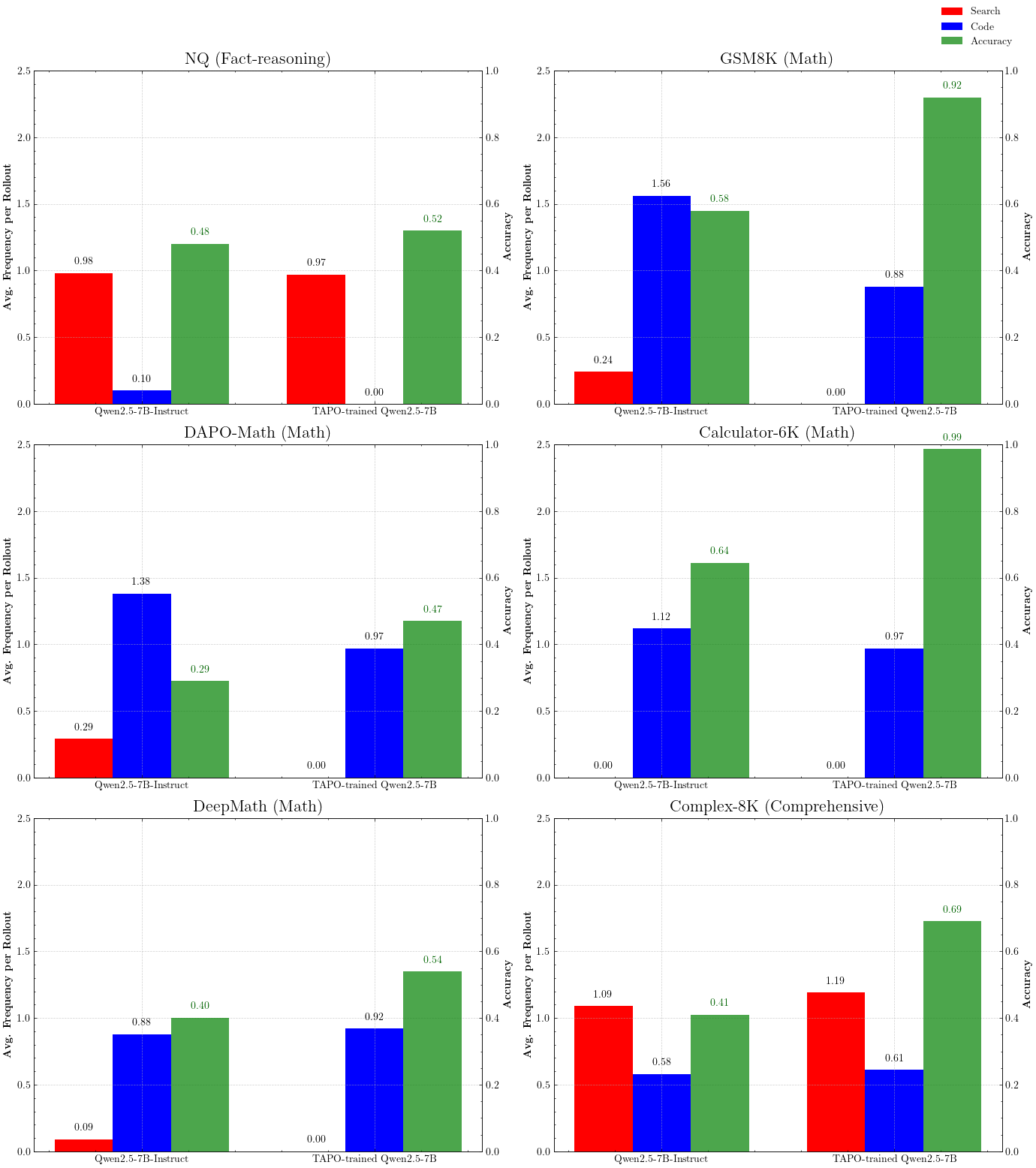}
\caption{Distribution of average tool invocation counts across benchmark datasets per rollout.}
\label{fig:tool_cnt}
\end{figure}

To evaluate the tool utilization efficiency of our approach, we conduct a quantitative analysis of tool invocation patterns in the 7B parameter model and the basic \textsc{Qwen2.5-7B}-Instruct model across diverse task categories. Figure~\ref{fig:tool_cnt} presents the comparative statistics of average tool invocation frequencies.

The results demonstrate several key behavioral patterns: the TAPO-trained model exhibits significantly better tool utilization compared to \textsc{Qwen2.5-7B}-Instruct. For knowledge-intensive tasks (e.g., NQ), the model exclusively activates the search engine tool; for mathematical reasoning tasks (e.g., GSM8K, DeepMath), it selectively employs only the code interpreter. Notably, in complex problem-solving scenarios (\texttt{complex-8K}), the model demonstrates the capability to dynamically interleave both tools when appropriate. This differentiated tool selection strategy indicates the model's ability to adapt its tool usage based on task requirements.

We conduct a comparative analysis of search engine invocation frequency between our model and Search-R1-7B on the NQ dataset. The results demonstrate comparable accuracy levels between both models, while Search-R1-7B exhibits significantly higher search frequency. Through qualitative case analysis, we identify instances of over-searching behavior in Search-R1-7B, where the model unnecessarily initiates search queries despite possessing adequate information to answer directly, as illustrated in our case studies (Appendix C 3).

\begin{figure}[ht]
\centering
\includegraphics[width=0.95\linewidth]{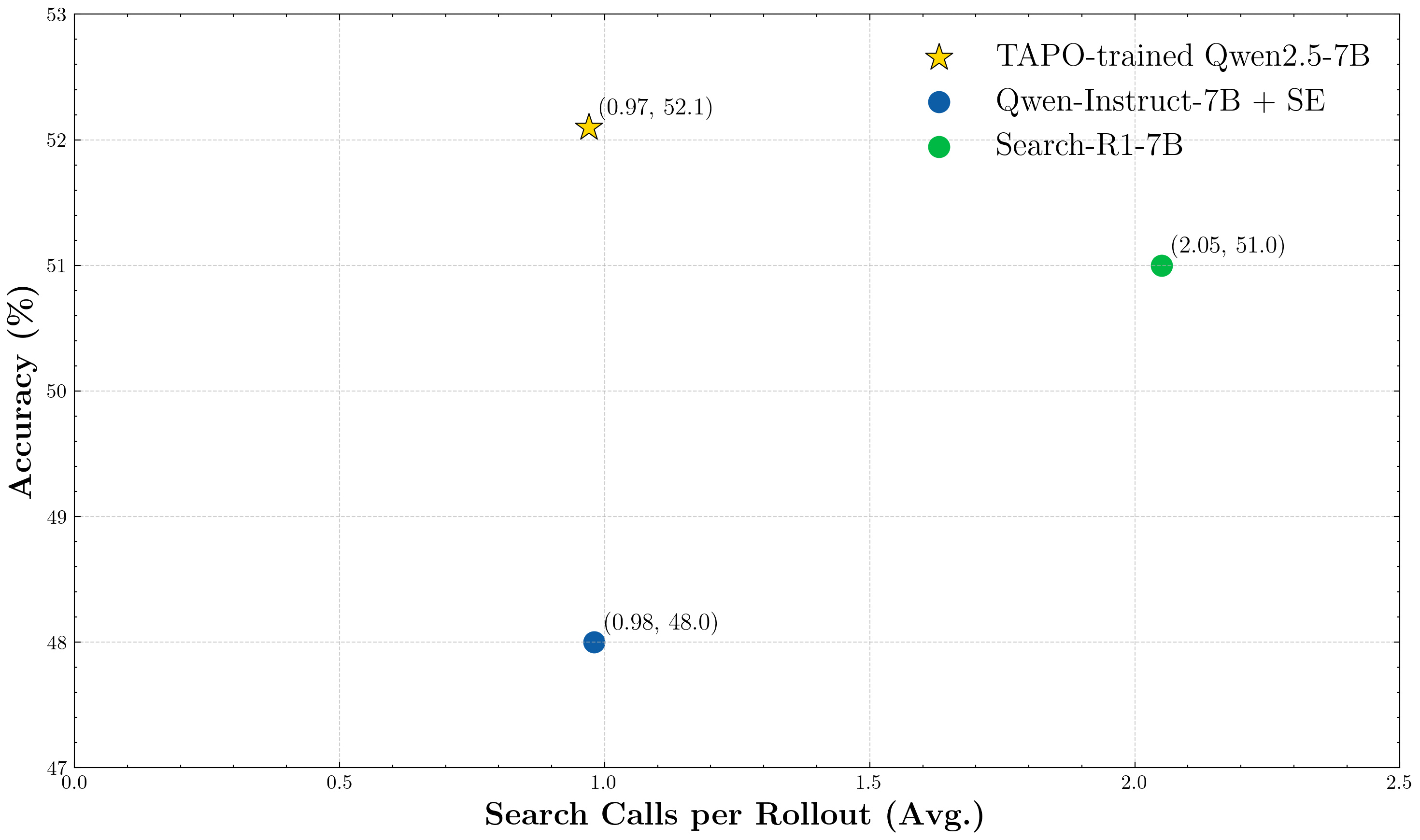}
\caption{Search engine invocation frequency versus performance on the NQ dataset.}
\label{fig:se_cnt}
\end{figure}


\section{Conclusions}
In this work, we introduced TAPO, a novel reinforcement learning framework that interleaves reasoning with adaptive search engine and code interpreter tool use in language models. We trained our 3B and 7B parameter models on two novel datasets (\texttt{TAPO-easy-60K} and \texttt{TAPO-hard-18K}) that require both knowledge retrieval and mathematical computation capabilities. Both models achieved strong performance and demonstrated high tool invocation efficiency, particularly in mathematical computation tasks.

Our ablation studies demonstrate that explicit reasoning steps substantially improve tool utilization efficiency in language models, particularly for code interpreter operations. Moreover, the TAPO-trained models exhibit robust resistance to over-invocation problems associated with reward hacking, owing to our carefully designed comprehensive dataset and strategically diversified toolset integration.


\bibliography{aaai2026}

\begin{thebibliography}{33}
\providecommand{\natexlab}[1]{#1}

\bibitem[{Ethayarajh et~al.(2024)Ethayarajh, Xu, Muennighoff, Jurafsky, and
  Kiela}]{ethayarajhKTOModelAlignment2024}
Ethayarajh, K.; Xu, W.; Muennighoff, N.; Jurafsky, D.; and Kiela, D. 2024.
\newblock {{KTO}}: {{Model Alignment}} as {{Prospect Theoretic Optimization}}.
\newblock arXiv:2402.01306.

\bibitem[{Feng et~al.(2025)Feng, Huang, Qu, Zhang, Qin, Zhong, Jiang, Chi, and
  Zhong}]{fengReToolReinforcementLearning2025}
Feng, J.; Huang, S.; Qu, X.; Zhang, G.; Qin, Y.; Zhong, B.; Jiang, C.; Chi, J.;
  and Zhong, W. 2025.
\newblock {{ReTool}}: {{Reinforcement Learning}} for {{Strategic Tool Use}} in
  {{LLMs}}.
\newblock arXiv:2504.11536.

\bibitem[{Gemini and DeepMind(2025)}]{deepmind2025gemini}
Gemini, G.; and DeepMind, G. 2025.
\newblock Gemini 2.5: Pushing the Frontier with Advanced Reasoning,
  Multimodality, Long Context, and Next Generation Agentic Capabilities.
\newblock Technical report, Google DeepMind.
\newblock Covers 3-hour video processing, MoE architecture, and agentic
  workflows.

\bibitem[{Google(2023)}]{serper2023}
Google. 2023.
\newblock Serper - Google Search API.
\newblock \url{https://serper.dev/}.
\newblock Fast and lightweight API for Google Search results with JSON output.

\bibitem[{Guo et~al.(2025)Guo, Yang, Zhang, Song, Zhang, Xu, Zhu, Ma, Wang, Bi
  et~al.}]{guo2025deepseek}
Guo, D.; Yang, D.; Zhang, H.; Song, J.; Zhang, R.; Xu, R.; Zhu, Q.; Ma, S.;
  Wang, P.; Bi, X.; et~al. 2025.
\newblock Deepseek-r1: Incentivizing reasoning capability in llms via
  reinforcement learning.
\newblock \emph{arXiv preprint arXiv:2501.12948}.

\bibitem[{Hendrycks et~al.(2021{\natexlab{a}})Hendrycks, Burns, Kadavath,
  Arora, Basart, Tang, Song, and Steinhardt}]{hendrycks2021measuring}
Hendrycks, D.; Burns, C.; Kadavath, S.; Arora, A.; Basart, S.; Tang, E.; Song,
  D.; and Steinhardt, J. 2021{\natexlab{a}}.
\newblock Measuring Mathematical Problem Solving With the {MATH} Dataset.
\newblock \emph{Advances in Neural Information Processing Systems}, 34:
  21828--21840.

\bibitem[{Hendrycks et~al.(2021{\natexlab{b}})Hendrycks, Burns, Kadavath,
  Arora, Basart, Tang, Song, and
  Steinhardt}]{hendrycks2021measuringmathematicalproblemsolving}
Hendrycks, D.; Burns, C.; Kadavath, S.; Arora, A.; Basart, S.; Tang, E.; Song,
  D.; and Steinhardt, J. 2021{\natexlab{b}}.
\newblock Measuring Mathematical Problem Solving With the MATH Dataset.
\newblock arXiv:2103.03874.

\bibitem[{Jaech et~al.(2024)Jaech, Kalai, Lerer, Richardson, El-Kishky, Low,
  Helyar, Madry, Beutel, Carney et~al.}]{jaech2024openai}
Jaech, A.; Kalai, A.; Lerer, A.; Richardson, A.; El-Kishky, A.; Low, A.;
  Helyar, A.; Madry, A.; Beutel, A.; Carney, A.; et~al. 2024.
\newblock Openai o1 system card.
\newblock \emph{arXiv preprint arXiv:2412.16720}.

\bibitem[{Jin et~al.(2025)Jin, Zeng, Yue, Yoon, Arik, Wang, Zamani, and
  Han}]{jin2025search}
Jin, B.; Zeng, H.; Yue, Z.; Yoon, J.; Arik, S.; Wang, D.; Zamani, H.; and Han,
  J. 2025.
\newblock Search-r1: Training llms to reason and leverage search engines with
  reinforcement learning.
\newblock \emph{arXiv preprint arXiv:2503.09516}.

\bibitem[{Joshi et~al.(2017)Joshi, Choi, Weld, and
  Zettlemoyer}]{joshi2017triviaqalargescaledistantly}
Joshi, M.; Choi, E.; Weld, D.~S.; and Zettlemoyer, L. 2017.
\newblock TriviaQA: A Large Scale Distantly Supervised Challenge Dataset for
  Reading Comprehension.
\newblock arXiv:1705.03551.

\bibitem[{Kaplan et~al.(2020)Kaplan, McCandlish, Henighan, Brown, Chess, Child,
  Gray, Radford, Wu, and Amodei}]{kaplanScalingLawsNeural2020}
Kaplan, J.; McCandlish, S.; Henighan, T.; Brown, T.~B.; Chess, B.; Child, R.;
  Gray, S.; Radford, A.; Wu, J.; and Amodei, D. 2020.
\newblock Scaling {{Laws}} for {{Neural Language Models}}.
\newblock arXiv:2001.08361.

\bibitem[{Kwiatkowski et~al.(2019)Kwiatkowski, Palomaki, Redfield, Collins,
  Parikh, Alberti, Epstein, Polosukhin, Devlin, Lee
  et~al.}]{kwiatkowski2019natural}
Kwiatkowski, T.; Palomaki, J.; Redfield, O.; Collins, M.; Parikh, A.; Alberti,
  C.; Epstein, D.; Polosukhin, I.; Devlin, J.; Lee, K.; et~al. 2019.
\newblock Natural questions: a benchmark for question answering research.
\newblock \emph{Transactions of the Association for Computational Linguistics},
  7: 453--466.

\bibitem[{Levenshtein et~al.(1966)}]{levenshtein1966binary}
Levenshtein, V.~I.; et~al. 1966.
\newblock Binary codes capable of correcting deletions, insertions, and
  reversals.
\newblock In \emph{Soviet physics doklady}, volume~10, 707--710. Soviet Union.

\bibitem[{Lewis et~al.(2020)Lewis, Perez, Piktus, Petroni, Karpukhin, Goyal,
  K{\"u}ttler, Lewis, Yih, Rockt{\"a}schel et~al.}]{lewis2020retrieval}
Lewis, P.; Perez, E.; Piktus, A.; Petroni, F.; Karpukhin, V.; Goyal, N.;
  K{\"u}ttler, H.; Lewis, M.; Yih, W.-t.; Rockt{\"a}schel, T.; et~al. 2020.
\newblock Retrieval-augmented generation for knowledge-intensive nlp tasks.
\newblock \emph{Advances in neural information processing systems}, 33:
  9459--9474.

\bibitem[{Lightman et~al.(2023)Lightman, Kosaraju, Burda, Edwards, Baker, Lee,
  Leike, Schulman, Sutskever, and Cobbe}]{lightman2023let}
Lightman, H.; Kosaraju, V.; Burda, Y.; Edwards, H.; Baker, B.; Lee, T.; Leike,
  J.; Schulman, J.; Sutskever, I.; and Cobbe, K. 2023.
\newblock Let's verify step by step.
\newblock In \emph{The Twelfth International Conference on Learning
  Representations}.

\bibitem[{Liu et~al.(2024)Liu, Feng, Xue, Wang, Wu, Lu, Zhao, Deng, Zhang, Ruan
  et~al.}]{liu2024deepseek}
Liu, A.; Feng, B.; Xue, B.; Wang, B.; Wu, B.; Lu, C.; Zhao, C.; Deng, C.;
  Zhang, C.; Ruan, C.; et~al. 2024.
\newblock Deepseek-v3 technical report.
\newblock \emph{arXiv preprint arXiv:2412.19437}.

\bibitem[{Meng, Xia, and Chen(2024)}]{meng2024simpo}
Meng, Y.; Xia, M.; and Chen, D. 2024.
\newblock Simpo: Simple preference optimization with a reference-free reward.
\newblock \emph{Advances in Neural Information Processing Systems}, 37:
  124198--124235.

\bibitem[{Nakano et~al.(2021)Nakano, Hilton, Balaji, Wu, Ouyang, Kim, Hesse,
  Jain, Kosaraju, Saunders et~al.}]{nakano2021webgpt}
Nakano, R.; Hilton, J.; Balaji, S.; Wu, J.; Ouyang, L.; Kim, C.; Hesse, C.;
  Jain, S.; Kosaraju, V.; Saunders, W.; et~al. 2021.
\newblock Webgpt: Browser-assisted question-answering with human feedback.
\newblock \emph{arXiv preprint arXiv:2112.09332}.

\bibitem[{OpenAI(2024)}]{hendrycks2024gpt4o}
OpenAI. 2024.
\newblock GPT-4o Technical Report.
\newblock \emph{OpenAI Blog}.
\newblock Model capabilities, safety, and multimodal performance details.

\bibitem[{Qwen et~al.(2025)Qwen, :, Yang, Yang, Zhang, Hui, Zheng, Yu, Li, Liu,
  Huang, Wei, Lin, Yang, Tu, Zhang, Yang, Yang, Zhou, Lin, Dang, Lu, Bao, Yang,
  Yu, Li, Xue, Zhang, Zhu, Men, Lin, Li, Tang, Xia, Ren, Ren, Fan, Su, Zhang,
  Wan, Liu, Cui, Zhang, and Qiu}]{qwen2024}
Qwen; :; Yang, A.; Yang, B.; Zhang, B.; Hui, B.; Zheng, B.; Yu, B.; Li, C.;
  Liu, D.; Huang, F.; Wei, H.; Lin, H.; Yang, J.; Tu, J.; Zhang, J.; Yang, J.;
  Yang, J.; Zhou, J.; Lin, J.; Dang, K.; Lu, K.; Bao, K.; Yang, K.; Yu, L.; Li,
  M.; Xue, M.; Zhang, P.; Zhu, Q.; Men, R.; Lin, R.; Li, T.; Tang, T.; Xia, T.;
  Ren, X.; Ren, X.; Fan, Y.; Su, Y.; Zhang, Y.; Wan, Y.; Liu, Y.; Cui, Z.;
  Zhang, Z.; and Qiu, Z. 2025.
\newblock Qwen2.5 Technical Report.
\newblock arXiv:2412.15115.

\bibitem[{Rafailov et~al.(2023)Rafailov, Sharma, Mitchell, Manning, Ermon, and
  Finn}]{rafailov2023direct}
Rafailov, R.; Sharma, A.; Mitchell, E.; Manning, C.~D.; Ermon, S.; and Finn, C.
  2023.
\newblock Direct preference optimization: Your language model is secretly a
  reward model.
\newblock \emph{Advances in Neural Information Processing Systems}, 36:
  53728--53741.

\bibitem[{Schick et~al.(2023)Schick, Dwivedi-Yu, Dess{\`\i}, Raileanu, Lomeli,
  Hambro, Zettlemoyer, Cancedda, and Scialom}]{schick2023toolformer}
Schick, T.; Dwivedi-Yu, J.; Dess{\`\i}, R.; Raileanu, R.; Lomeli, M.; Hambro,
  E.; Zettlemoyer, L.; Cancedda, N.; and Scialom, T. 2023.
\newblock Toolformer: Language models can teach themselves to use tools.
\newblock \emph{Advances in Neural Information Processing Systems}, 36:
  68539--68551.

\bibitem[{Schulman et~al.(2017)Schulman, Wolski, Dhariwal, Radford, and
  Klimov}]{schulman2017proximal}
Schulman, J.; Wolski, F.; Dhariwal, P.; Radford, A.; and Klimov, O. 2017.
\newblock Proximal policy optimization algorithms.
\newblock \emph{arXiv preprint arXiv:1707.06347}.

\bibitem[{Shao et~al.(2024)Shao, Wang, Zhu, Xu, Song, Bi, Zhang, Zhang, Li, Wu
  et~al.}]{shao2024deepseekmath}
Shao, Z.; Wang, P.; Zhu, Q.; Xu, R.; Song, J.; Bi, X.; Zhang, H.; Zhang, M.;
  Li, Y.; Wu, Y.; et~al. 2024.
\newblock Deepseekmath: Pushing the limits of mathematical reasoning in open
  language models.
\newblock \emph{arXiv preprint arXiv:2402.03300}.

\bibitem[{Snell et~al.(2024)Snell, Lee, Xu, and
  Kumar}]{snellScalingLLMTestTime2024}
Snell, C.; Lee, J.; Xu, K.; and Kumar, A. 2024.
\newblock Scaling {{LLM Test-Time Compute Optimally}} Can Be {{More Effective}}
  than {{Scaling Model Parameters}}.
\newblock arXiv:2408.03314.

\bibitem[{Vaswani et~al.(2017)Vaswani, Shazeer, Parmar, Uszkoreit, Jones,
  Gomez, Kaiser, and Polosukhin}]{vaswani2017attention}
Vaswani, A.; Shazeer, N.; Parmar, N.; Uszkoreit, J.; Jones, L.; Gomez, A.~N.;
  Kaiser, {\L}.; and Polosukhin, I. 2017.
\newblock Attention is all you need.
\newblock \emph{Advances in neural information processing systems}, 30.

\bibitem[{Wei et~al.(2022)Wei, Wang, Schuurmans, Bosma, Xia, Chi, Le, Zhou
  et~al.}]{wei2022chain}
Wei, J.; Wang, X.; Schuurmans, D.; Bosma, M.; Xia, F.; Chi, E.; Le, Q.~V.;
  Zhou, D.; et~al. 2022.
\newblock Chain-of-thought prompting elicits reasoning in large language
  models.
\newblock \emph{Advances in neural information processing systems}, 35:
  24824--24837.

\bibitem[{Wu et~al.(2025{\natexlab{a}})Wu, Li, Fang, Yin, Zhang, Tao, Zhang,
  Xi, Jiang, Xie et~al.}]{wu2025webdancer}
Wu, J.; Li, B.; Fang, R.; Yin, W.; Zhang, L.; Tao, Z.; Zhang, D.; Xi, Z.;
  Jiang, Y.; Xie, P.; et~al. 2025{\natexlab{a}}.
\newblock WebDancer: Towards Autonomous Information Seeking Agency.
\newblock \emph{arXiv preprint arXiv:2505.22648}.

\bibitem[{Wu et~al.(2025{\natexlab{b}})Wu, Sun, Li, Welleck, and
  Yang}]{wuINFERENCESCALINGLAWS2025}
Wu, Y.; Sun, Z.; Li, S.; Welleck, S.; and Yang, Y. 2025{\natexlab{b}}.
\newblock {{Inference scaling laws}}: {{an empirical analysis of
  compute-optimal inference for llm problem-solving}}.
\newblock \emph{Conference on Neural Information Processing Systems}.

\bibitem[{Yang et~al.(2018)Yang, Qi, Zhang, Bengio, Cohen, Salakhutdinov, and
  Manning}]{yang2018hotpotqadatasetdiverseexplainable}
Yang, Z.; Qi, P.; Zhang, S.; Bengio, Y.; Cohen, W.~W.; Salakhutdinov, R.; and
  Manning, C.~D. 2018.
\newblock HotpotQA: A Dataset for Diverse, Explainable Multi-hop Question
  Answering.
\newblock arXiv:1809.09600.

\bibitem[{Yao et~al.(2023)Yao, Zhao, Yu, Du, Shafran, Narasimhan, and
  Cao}]{yaoReActSynergizingReasoning2023}
Yao, S.; Zhao, J.; Yu, D.; Du, N.; Shafran, I.; Narasimhan, K.; and Cao, Y.
  2023.
\newblock {{ReAct}}: {{Synergizing Reasoning}} and {{Acting}} in {{Language
  Models}}.
\newblock arXiv:2210.03629.

\bibitem[{Yu et~al.(2025)Yu, Zhang, Zhu, Yuan, Zuo, Yue, Dai, Fan, Liu, Liu
  et~al.}]{yu2025dapo}
Yu, Q.; Zhang, Z.; Zhu, R.; Yuan, Y.; Zuo, X.; Yue, Y.; Dai, W.; Fan, T.; Liu,
  G.; Liu, L.; et~al. 2025.
\newblock Dapo: An open-source llm reinforcement learning system at scale.
\newblock \emph{arXiv preprint arXiv:2503.14476}.

\bibitem[{Zhou et~al.(2023)Zhou, Yan, Shlapentokh-Rothman, Wang, and
  Wang}]{zhou2023language}
Zhou, A.; Yan, K.; Shlapentokh-Rothman, M.; Wang, H.; and Wang, Y.-X. 2023.
\newblock Language agent tree search unifies reasoning acting and planning in
  language models.
\newblock \emph{arXiv preprint arXiv:2310.04406}.

\end{thebibliography}

\clearpage
\onecolumn

\begin{appendices}

\section{Dataset Specifications}
\label{app:dataset}
To rigorously evaluate our reinforcement learning framework TAPO, we curated two specialized question-answering datasets: (1) \texttt{TAPO-easy-60K} and (2) \texttt{TAPO-hard-18K}. These datasets encompass both mathematical problem-solving tasks and fact-based reasoning challenges. Crucially, we designed domain-specific reward functions for each data source to ensure precise and meaningful feedback signals during training.

\begin{table}[htbp]
\centering
\small
\caption{Datasets and Reward Functions in \texttt{TAPO-easy-60K}, where $\text{lev}(a,b)$ denotes the Levenshtein distance (minimum edit operations via insertion, deletion, or substitution required to transform string $a$ into $b$)}
\label{tab:easy}
\begin{tabular}{>{\raggedright}p{3cm}>{\raggedright}p{2.5cm}c>{\raggedright\arraybackslash}p{7cm}}
\toprule
\textbf{Dataset} & \textbf{Type} & \textbf{Size} & \textbf{Reward Function} \\
\midrule
NQ & Fact reasoning & 33K & 
$f(o, \text{gt}) = \Bigl(1 - r(o, \text{gt})\Bigr) \cdot \mathbb{I}\Bigl[r(o, \text{gt}) < 0.5\Bigr]$ \\
& & & $r(o, \text{gt}) = \dfrac{\text{lev}(o, \text{gt})}{\max(|o|, |\text{gt}|)}$ \\
\addlinespace
Calculator-6K & Math & 6K & 
$f(o, \text{gt}) = \mathbb{I} \Bigl[|o - \text{gt}| \le 0.001 \Bigr]$ \\
\addlinespace
GSM8K & Math & 8K & 
$f(o, \text{gt}) = \mathbb{I}\Bigl[o \equiv \text{gt}\Bigr]$ \\
\addlinespace
DAPO-Math & Math & 13K & 
$f(o, \text{gt}) = \mathbb{I}\Bigl[o \equiv \text{gt}\Bigr]$ \\
\bottomrule
\end{tabular}
\end{table}

\begin{table}[htbp]
\centering
\small
\caption{Datasets and Their Reward Functions in \texttt{TAPO-hard-18K}.}
\label{tab:hard}
\begin{tabular}{>{\raggedright}p{3cm}>{\raggedright}p{2.5cm}c>{\raggedright\arraybackslash}p{7cm}}
\toprule
\textbf{Dataset} & \textbf{Type} & \textbf{Size} & \textbf{Reward Function} \\
\midrule
DeepMath & Math & 10K & 
$f(o, \text{gt}) = \mathbb{I}\Bigl[o \equiv \text{gt}\Bigr]$ \\
\addlinespace
Complex-8K & Comprehensive & 8K &
$f(o, \text{gt}) = \mathbb{I}\Bigl[|o - \text{gt}| \le 0.005 |\text{gt}|\Bigr]$ \\
\bottomrule
\end{tabular}
\end{table}

It is particularly noteworthy that the datasets Calculator-6K and Complex-8K were developed specifically for this study. Calculator-6K is a question-answering dataset containing approximately 6,000 simple mathematical calculation problems, generated automatically using Python scripts with predefined templates. Complex-8K represents a more comprehensive dataset that requires both retrieval capabilities for accessing external knowledge and mathematical computation skills. Below are representative examples:

\begin{table}[htbp]
\centering
\small
\caption{Examples from Calculator-6K, featuring computationally intensive problems that encourage code interpreter usage.}
\label{tab:calculator}
\begin{tabular}{l|l}
\toprule
\textbf{Question} & \textbf{Ground Truth} \\
\midrule
What is -16008947 divided by 86 plus -11867948.017539188 times -33830597.53172473? & 401499772722648.062500 \\
\addlinespace
What is the Euclidean distance between points (-59.1, -84) and (521, 8)? & 587.349989 \\
\addlinespace
What is the 3D Euclidean distance between (-86, -63, -43) and (-77, 49, -24)? & 113.956131 \\
\addlinespace
What is the square of 34.31096597945637? & 1177.242386 \\
\addlinespace
What is -34053513.64467979 * 1963819 - 47599034? & -66874984711215.421875 \\
\addlinespace
What's the sum of 60.61620772833686, 3332093.3462560773 and 18071071? & 21403224.962464 \\
\addlinespace
What is 10842057.978939041 / -20 + -54282778.56760053? & -54824881.466547 \\
\addlinespace
How much is 55226286 plus 8692820.413110033? & 63919106.413110 \\
\addlinespace
What is the Manhattan distance between points (-72614440.6, 48577125) and (48, 76.1)? & 121191689.700000 \\
\addlinespace
What is the sine of 75203372.15922138 radians? & 0.501888 \\
\bottomrule
\end{tabular}
\end{table}

\begin{table}[htbp]
\centering
\small
\caption{Examples from Complex-8K, require external knowledge retrieval and simple code-based computation, offering comprehensive evaluation.}
\label{tab:complex}
\begin{tabular}{p{0.7\linewidth}|p{0.15\linewidth}}
\toprule
\textbf{Question} & \textbf{Ground Truth} \\
\midrule
How many liters of beer were consumed per capita in Germany in 2020? (Unit: liters, round to one decimal place) & 92.4 \\
\addlinespace
What was the volume of international container traffic through the Port of Shanghai in 2023? (Unit: million TEUs, round to nearest integer) & 49.0 \\
\addlinespace
Calculate the average daily stock price of Apple Inc. (AAPL) during December 2020. (Unit: US dollars, round to two decimal places) & 129.42 \\
\addlinespace
Determine the total investment in renewable energy projects in Australia in 2022. (Unit: billion Australian dollars, round to one decimal place) & 6.2 \\
\addlinespace
What was the closing value of the Dow Jones Industrial Average on December 31, 2023? (Unit: index points) & 37689.54 \\
\addlinespace
How much did the market cap of Apple change in 2023? (Unit: US dollars, round to nearest billion) & 700.0 \\
\addlinespace
What was the average CO2 concentration in the atmosphere in 2021? (Unit: parts per million, round to one decimal place) & 416.3 \\
\addlinespace
How many solar panels were installed in Spain from 2020 to 2024? (Unit: panels) & 1942.0 \\
\addlinespace
How many kilometers does the average person in Japan travel by train per year between 2020 and 2024? (Unit: kilometers, round to nearest integer) & 45.33 \\
\addlinespace
How many new startups were founded in Silicon Valley during the year 2023? (Unit: startups) & 160 \\
\addlinespace
What is the average number of active satellites in Earth's orbit as of 2022? (Unit: satellites, no decimal places) & 5785 \\
\addlinespace
Compute the GDP growth rate of India from the fiscal year 2021 to 2024. (Unit: percentage, round to one decimal place) & 23.5 \\
\addlinespace
Calculate the growth in electric vehicle sales in China between 2021 and 2023. (Unit: percentage, round to one decimal place) & 523.1 \\
\addlinespace
What was the average monthly unemployment rate in the Eurozone for the year 2020? (Unit: percentage, round to one decimal place) & 5.4 \\
\addlinespace
How much did the S\&P 500 index grow between January 2021 and January 2024? (Unit: index points, round to one decimal place) & 1097.2 \\
\addlinespace
What was the average length of a YouTube video uploaded in 2023? (Unit: minutes, round to two decimal places) & 12.43 \\
\addlinespace
What was the total number of tourists visiting Hawaii in 2022? (Unit: tourists, round to nearest hundred thousand) & 9200000 \\
\addlinespace
What was the inflation rate in Germany for the entire year of 2022? (Unit: percentage, round to one decimal place) & 7.9 \\
\addlinespace
Calculate the compound annual growth rate (CAGR) of Apple's revenue from 2020 to 2023. (Unit: percentage, round to two decimal places) & 11.77 \\
\addlinespace
What was the total carbon emissions from the aviation sector in the UK in 2020? (Unit: million tons CO2, round to one decimal place) & 16.4 \\
\addlinespace
How many new species were discovered in the Amazon rainforest from 2020 to 2023? (Unit: species) & 441 \\
\addlinespace
Calculate the length of time between the SpaceX Starship first launch and the Artemis I mission launch. (Unit: days) & 155 \\
\bottomrule
\end{tabular}
\end{table}

\section{Supplementary Experimental Results and Analysis}
\label{app:exp}
In this section, we present additional details about our experiments.

\subsection{Experimental Setup Details}
The project was implemented using the VeRL reinforcement learning framework. All experiments were conducted on \textsc{Qwen2.5-3B} and \textsc{Qwen2.5-7B} models, and the code will be made publicly available soon. The following studies focus on the 7B model training.

\subsection{Reward Analysis}

\begin{figure}[H]
    \centering
    \includegraphics[width=0.8\linewidth]{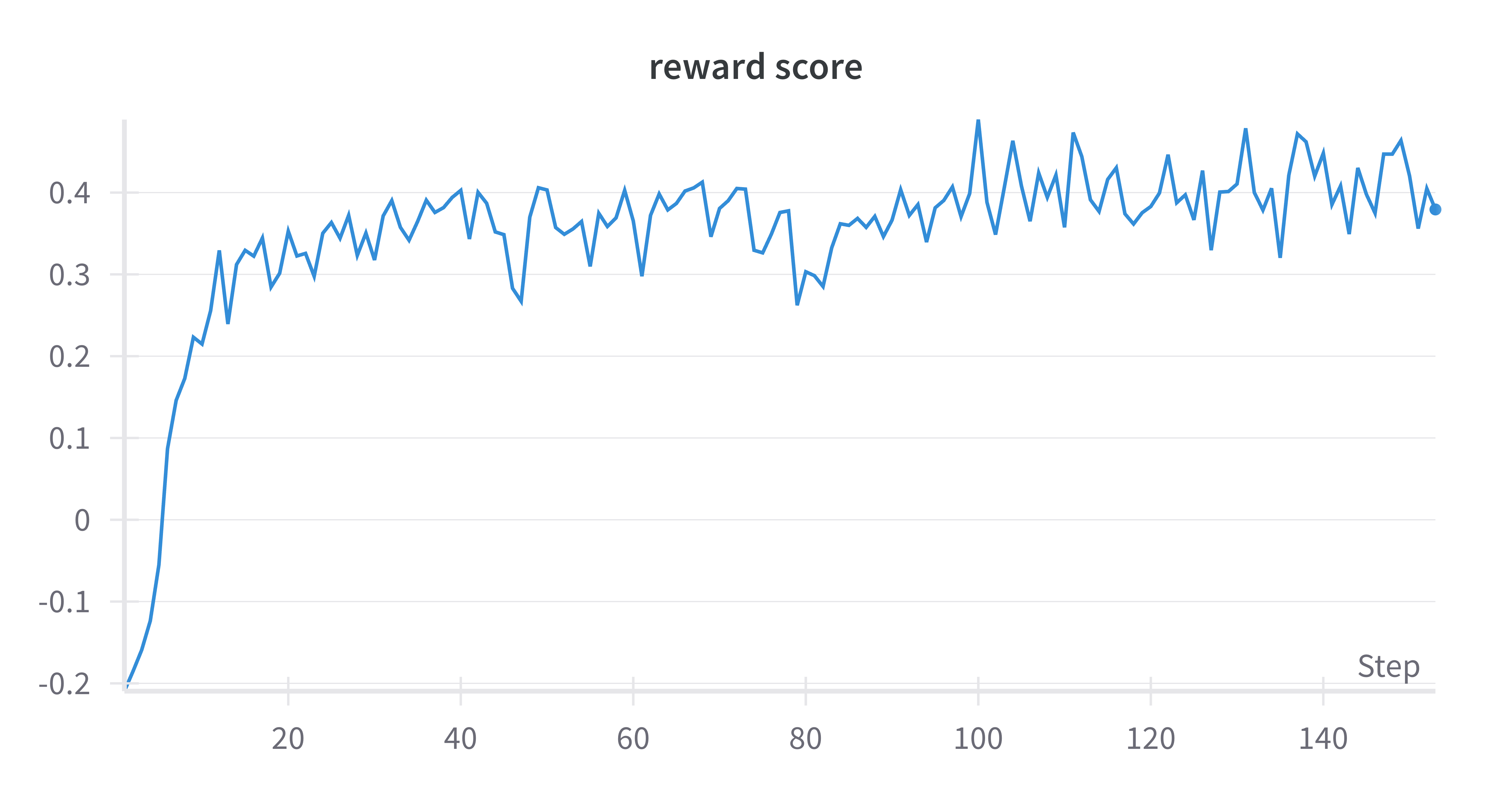}
    \caption{Reward score progression during training.}
    \label{fig:reward}
\end{figure}

\subsection{Entropy Analysis}
Although the policy model achieved strong performance on multiple benchmarks, we observed a sharp entropy decline from approximately 1.6 to 0.2. This performance is notably worse than standard DAPO training, which maintains an entropy level around 0.6.

This finding suggests that tool invocation in reinforcement learning may negatively impact entropy, potentially leading to less diverse model outputs.

\begin{figure}[H]
    \centering
    \includegraphics[width=0.8\linewidth]{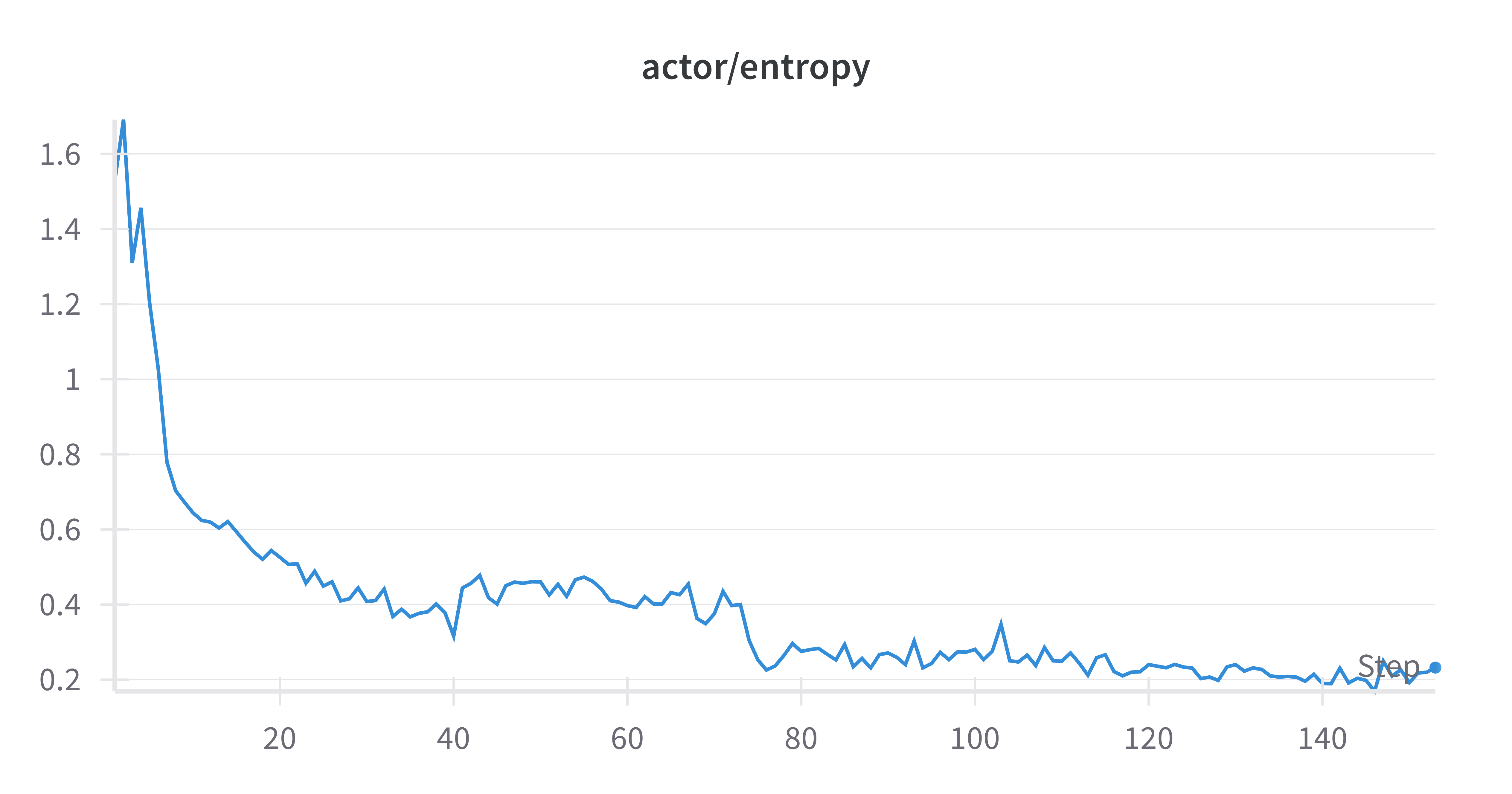}
    \caption{Entropy progression during training.}
    \label{fig:entropy}
\end{figure}

\subsection{Response Length Analysis}
During training, we observed a significant decrease in the policy model's response length. This indicates that the model learned to generate concise yet informative intermediate steps rather than producing lengthy responses to obtain correct answers.

\begin{figure}[H]
    \centering
    \includegraphics[width=0.8\linewidth]{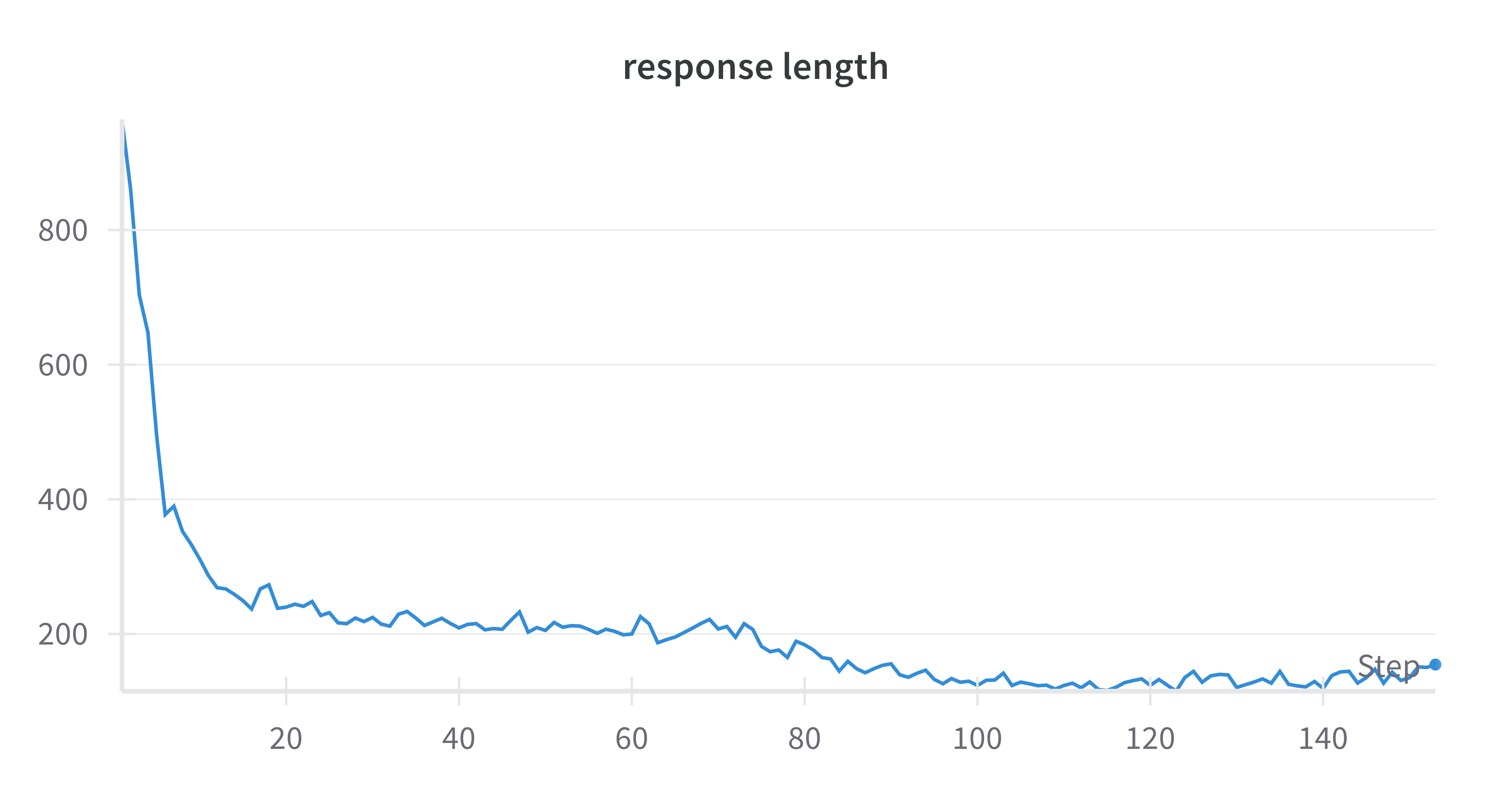}
    \caption{Response length progression during training.}
    \label{fig:response}
\end{figure}

\section{Evaluation Case Studies}
\label{app:eval}
In this section, we present several case studies demonstrating our method. For clarity, our TAPO-trained Qwen2.5-7B model will be referred to simply as TAPO-7B throughout these examples.

\subsection{Case 1}

Standard language models relying solely on direct inference frequently exhibit hallucination issues and lack access to current information, whereas TAPO-7B effectively addresses these limitations through integrated search engine and code interpreter capabilities.

\definecolor{bad}{RGB}{255, 230, 230}
\definecolor{good}{RGB}{230, 240, 255}

\subsubsection{Case 1.1}
\textbf{Question: Which number is larger, 9.11 or 9.8?}
\begin{tcbraster}[
    raster columns=1,
    enhanced,   
    sharp corners
]

\begin{tcolorbox}[
    title=Qwen2.5-Instruct-7B + CoT,
    colback=bad,
    colframe=red!60,
    fonttitle=\bfseries,
    title style={fill=red!20}
]
\texttt{<think>}When comparing decimal numbers, we start by comparing the whole number parts first. Both numbers have the same whole number part, which is 9. Next, we compare the decimal parts. The first number has a decimal part of .11 and the second number has a decimal part of .8. Since 11 is greater than 8, .11 is greater than .8. Therefore, 9.11 is larger than 9.8.\texttt{</think>}

\texttt{<answer>}9.11\texttt{</answer>} \wrong{}
\end{tcolorbox}

\begin{tcolorbox}[
    title=TAPO-7B,
    colback=good,
    colframe=blue!60,
    fonttitle=\bfseries,
    title style={fill=blue!20}
]
\texttt{<code>} \\
\texttt{print(9.8 > 9.11)} \\
\texttt{</code>}

\texttt{<response>}True \\
\texttt{</response>}

\texttt{<answer>}9.8\texttt{</answer>} \correct{}
\end{tcolorbox}

\end{tcbraster}

\subsubsection{Case 1.2}
\textbf{Question: How many 'r's are in the word 'strawberry'?}
\begin{tcbraster}[
    raster columns=1,
    enhanced,   
    sharp corners
]

\begin{tcolorbox}[
    title=Qwen2.5-Instruct-7B + CoT,
    colback=bad,
    colframe=red!60,
    fonttitle=\bfseries,
    title style={fill=red!20}
]
\texttt{<think>}Let's count the 'r's in the word 'strawberry'. The word is composed of the following letters: s-t-r-a-w-b-e-r-r-y. We can see that there are two 'r's in this word.\texttt{</think>}

\texttt{<answer>}2\texttt{</answer>} \wrong{}
\end{tcolorbox}

\begin{tcolorbox}[
    title=TAPO-7B,
    colback=good,
    colframe=blue!60,
    fonttitle=\bfseries,
    title style={fill=blue!20}
]
\texttt{<code>} \\
\texttt{word = "strawberry" \\
count = word.count('r') \\
print(count)} \\
\texttt{</code>}

\texttt{<response>}3 \\
\texttt{</response>}

\texttt{<answer>}3\texttt{</answer>} \correct{}
\end{tcolorbox}

\end{tcbraster}

\subsubsection{Case 1.3}
\textbf{Question: Who is the current U.S. president?}
\begin{tcbraster}[
    raster columns=1,
    enhanced,   
    sharp corners
]

\begin{tcolorbox}[
    title=Qwen2.5-Instruct-7B + CoT,
    colback=bad,
    colframe=red!60,
    fonttitle=\bfseries,
    title style={fill=red!20}
]
\texttt{<think>}As of 2023, Joe Biden is serving as the 46th president of the United States.\texttt{</think>}

\texttt{<answer>}Joe Biden\texttt{</answer>} \wrong{}
\end{tcolorbox}

\begin{tcolorbox}[
    title=TAPO-7B,
    colback=good,
    colframe=blue!60,
    fonttitle=\bfseries,
    title style={fill=blue!20}
]
\texttt{<think>}
To answer this question, I need to search for the current U.S. president.
\texttt{</think>}

\texttt{<search>}current U.S. president\texttt{</search>}

\texttt{<response>}

Doc 1: (Knowledge Panel: Donald Trump) Donald John Trump is an American politician, media personality, and businessman who is the 47th president of the United States. A member of the Republican Party, he served as the 45th president from 2017 to 2021.

Doc 2: (Title: President of the United States - Wikipedia) In all, 45 individuals have served 47 presidencies spanning 60 four-year terms. Donald Trump is the 47th and current president since January 20, 2025.

Doc 3: (Title: The White House) Donald J. Trump. President of the United States · JD Vance. VICE PRESIDENT OF THE UNITED STATES · Melania Trump. First Lady OF THE UNITED STATES · The Cabinet. Of ...

Doc 4: (Title: Presidents, vice presidents, and first ladies | USAGov) The 47th and current president of the United States is Donald John Trump. He was sworn into office on January 20, 2025.

Doc 5: (Title: The Trump Administration - The White House) After a landslide election victory in 2024, President Donald J. Trump is returning to the White House to build upon his previous successes and use his mandate ...

Doc 6: (Title: List of presidents of the United States - Wikipedia) The incumbent president is Donald Trump, who assumed office on January 20, 2025. Since the office was established in 1789, 45 men have served ...

\texttt{</response>}

\texttt{<answer>}Donald Trump\texttt{</answer>} \correct{}

\end{tcolorbox}

\end{tcbraster}

\subsection{Case 2}

This section presents two representative examples from the Complex-8K dataset, demonstrating how multi-stage reasoning and reinforcement learning enhances the model's tool utilization capabilities.

\subsubsection{Case 2.1}
\textbf{Question: Calculate the length of time between the SpaceX Starship first launch and the Artemis I mission launch. (Unit: days)}
\begin{tcbraster}[
    raster columns=1,
    enhanced,   
    sharp corners
]

\begin{tcolorbox}[
    title=Qwen2.5-Instruct-7B + SE/CI,
    colback=bad,
    colframe=red!60,
    fonttitle=\bfseries,
    title style={fill=red!20}
]
\texttt{<think>}To solve this problem, I need to find the dates of the first SpaceX Starship launch and the Artemis I mission launch, then calculate the difference in days between these two dates.\texttt{</think>}

\texttt{<search>}SpaceX Starship first launch date and Artemis I mission launch date\texttt{</search>}

\texttt{<response>}

Doc 1: (Title: List of Starship launches - Wikipedia) ... mission slated for launch to the moon in 2027 as part of SpaceX's first Starship HLS mission for the Artemis program. An unknown number of refueling ...

Doc 2: (Title: SpaceX Starship - Wikipedia) The first launch of a full Starship vehicle occurred on April 20, 2023, and ended with the explosion of the rocket four minutes after liftoff. The program has ...

Doc 3: (Title: Artemis I - NASA) Launch date: Nov. 16, 2022 ; Mission duration: 25 days, 10 hours, 53 minutes ; Total distance traveled: 1.4 million miles ; Re-entry speed: 24,581 mph (Mach 32).

Doc 4: (Title: Starship overview - SpaceX) Starship is the world's most powerful launch vehicle ever developed, capable of carrying up to 150 metric tonnes fully reusable and 250 metric tonnes expendable ...

Doc 5: (Title: NASA's Artemis Moon Missions: all you need to know) Artemis 1: After being delayed four times, Artemis 1 successfully launched on Wednesday 16 November 2022 at 1:47am EST (6:47am GMT); Artemis 2: No earlier than ...

\texttt{</response>}

\texttt{<answer>}37\texttt{</answer>} \wrong{}
\end{tcolorbox}

\begin{tcolorbox}[
    title=TAPO-7B,
    colback=good,
    colframe=blue!60,
    fonttitle=\bfseries,
    title style={fill=blue!20}
]
\texttt{<think>}To find the days between the first SpaceX Starship launch and Artemis I, I need their dates and calculate the difference.\texttt{</think>}

\texttt{<search>}SpaceX Starship first launch date and Artemis I mission launch date\texttt{</search>}

\texttt{<response>}

Doc 1: (Title: List of Starship launches - Wikipedia) ... mission slated for launch to the moon in 2027 as part of SpaceX's first Starship HLS mission for the Artemis program. An unknown number of refueling ...

Doc 2: (Title: SpaceX Starship - Wikipedia) The first launch of a full Starship vehicle occurred on April 20, 2023, and ended with the explosion of the rocket four minutes after liftoff. The program has ...

Doc 3: (Title: Artemis I - NASA) Launch date: Nov. 16, 2022 ; Mission duration: 25 days, 10 hours, 53 minutes ; Total distance traveled: 1.4 million miles ; Re-entry speed: 24,581 mph (Mach 32).

Doc 4: (Title: Starship overview - SpaceX) Starship is the world's most powerful launch vehicle ever developed, capable of carrying up to 150 metric tonnes fully reusable and 250 metric tonnes expendable ...

Doc 5: (Title: NASA's Artemis Moon Missions: all you need to know) Artemis 1: After being delayed four times, Artemis 1 successfully launched on Wednesday 16 November 2022 at 1:47am EST (6:47am GMT); Artemis 2: No earlier than ...

\texttt{</response>}

\texttt{<think>}Starship's first launch was on 2023-04-20, and Artemis I launched on 2022-11-16. I'll calculate the difference using Python datetime.\texttt{</think>}

\texttt{<code>} \\
\texttt{import datetime \\
first\_launch = datetime.datetime(2023, 4, 20) \\
artemis\_i = datetime.datetime(2022, 11, 16) \\
days = (artemis\_i - first\_launch).days \\
print(days) \\}
\texttt{</code>}

\texttt{<response>}-155 \\
\texttt{</response>}

\texttt{<answer>}155\texttt{</answer>} \correct{}
\end{tcolorbox}

\end{tcbraster}

\subsubsection{Case 2.2}
\textbf{Question: Compute the GDP growth rate of India from the fiscal year 2021 to 2024. (Unit: percentage, round to one decimal place)}
\begin{tcbraster}[
    raster columns=1,
    enhanced,   
    sharp corners
]

\begin{tcolorbox}[
    title=Qwen2.5-Instruct-7B + SE/CI,
    colback=bad,
    colframe=red!60,
    fonttitle=\bfseries,
    title style={fill=red!20}
]
\texttt{<think>}
To compute the GDP growth rate from fiscal year 2021 to 2024 for India, I need to gather the GDP values for these years. Then, I can use the formula for calculating the growth rate:

\[ \text{Growth Rate} = \left( \frac{\text{GDP}_{\text{Year 2}} - \text{GDP}_{\text{Year 1}}}{\text{GDP}_{\text{Year 1}}} \right) \times 100 \]

I will assume that the GDP figures are in constant prices and in billions of local currency units for simplicity, as the exact currency is not specified.
\texttt{</think>}

\texttt{<search>}GDP of India 2021-2024\texttt{</search>}

\texttt{<response>}

Doc 1: (Title: Economy of India - Wikipedia) Economy of India ; –6.4\% of GDP (2022–23) · 34.96 trillion (US\$410 billion); (2025–26) · 50.65 trillion (US\$600 billion); (2025–26) · Donor: Increase \$4.234 ...

Doc 2: (Title: GDP (current US\$) - India - World Bank Data) GDP (current US\$) - India. Country official statistics, National Statistical Organizations and/or Central Banks; National Accounts data files, Organisation ...

Doc 3: (Title: India GDP | Historical Chart \& Data - Macrotrends) India GDP for 2022 was 3.353 trillion US dollars, a 5.88\% increase from 2021. · India GDP for 2021 was 3.167 trillion US dollars, a 18.41\% increase from 2020.

Doc 4: (Title: India Fiscal Year GDP Growth - Trading Economics) Full Year GDP Growth in India decreased to 6.50 percent in 2025 from 9.20 percent in 2024. Full Year GDP Growth in India averaged 6.42 percent from 2006 until ...

Doc 5: (Title: India - IMF DataMapper) Real GDP growth. Annual percent change. 6.2 ; GDP, current prices. Billions of U.S. dollars. 4.19 ; GDP per capita, current prices. U.S. dollars per capita. 2.88.

\texttt{</response>}

\texttt{<code>} \\
\texttt{\# Define the GDP values in trillions of Indian Rupees} \\
\texttt{gdp\_2021 = 31.67} \\
\hl{\texttt{gdp\_2024 = 41.00}} \textcolor{red!70!black}{(Hallucination !)} \\

\texttt{\# Calculate the growth rate} \\
\texttt{growth\_rate = ((gdp\_2024 - gdp\_2021) / gdp\_2021) * 100} \\
\texttt{print(f"{growth\_rate:.1f}\%")} \\
\texttt{</code>}

\texttt{<response>}29.5\% \\
\texttt{</response>}

\texttt{<answer>}29.5\texttt{</answer>} \wrong{}
\end{tcolorbox}

\begin{tcolorbox}[
    title=TAPO-7B,
    colback=good,
    colframe=blue!60,
    fonttitle=\bfseries,
    title style={fill=blue!20}
]
\texttt{<think>}I need to find the GDP values for India for the years 2021 and 2024. Then, I will compute the growth rate using these values. Let's search for the GDP data.\texttt{</think>}

\texttt{<search>}GDP of India 2021 and 2024\texttt{</search>}

\texttt{<response>}

Doc 1: (Title: Economy of India - Wikipedia) Increase \$514 billion (2021–22) · Negative increase –\$44.563 billion (2024) · Negative increase –1.146\% of GDP (2024) · Negative increase \$663.8 billion (March ...

Doc 2: (Title: India GDP | Historical Chart \& Data - Macrotrends) India GDP for 2022 was 3.353 trillion US dollars, a 5.88\% increase from 2021. · India GDP for 2021 was 3.167 trillion US dollars, a 18.41\% increase from 2020.

Doc 3: (Title: GDP growth (annual \%) - India - World Bank Data) 2024. 3.6. Hong Kong SAR, China. 2024. 2.5. Hungary. 2024. 0.5. Iceland. 2024. 0.5. India. 2024. 6.5. Indonesia. 2024. 5.0. Iran, Islamic Rep. 2024. 3.0. Iraq.

Doc 4: (Title: GDP of India 1987-2030 - Statista) In 2024, GDP in India was at around 3.91 trillion US dollars, and it is expected to reach six trillion by the end of the decade.

Doc 5: (Title: India Fiscal Year GDP Growth - Trading Economics) Full Year GDP Growth in India decreased to 6.50 percent in 2025 from 9.20 percent in 2024. Full Year GDP Growth in India is expected to reach 6.30 percent by ...

\texttt{</response>}

\texttt{<think>}
Based on the provided information, I have the GDP figures for India for the fiscal years 2021 and 2024.

For 2021:

- GDP: 3.167 trillion

For 2024:

- GDP: 3.91 trillion
\texttt{</think>}

\texttt{<code>} \\
\texttt{gdp\_2021 = 3.167e12 \# in USD} \\
\texttt{gdp\_2024 = 3.91e12   \# in USD} \\

\texttt{growth\_rate = ((gdp\_2024 - gdp\_2021) / gdp\_2021) * 100} \\
\texttt{print(round(growth\_rate, 1))} \\
\texttt{</code>}

\texttt{<response>}23.5 \\
\texttt{</response>}

\texttt{<answer>}23.5\texttt{</answer>} \correct{}
\end{tcolorbox}

\end{tcbraster}

\subsection{Case 3}

While Search-R1-7B and TAPO-7B demonstrate comparable accuracy on the NQ dataset, TAPO-7B achieves significantly higher tool invocation efficiency by requiring fewer search operations. Below we analyze two representative NQ examples illustrating this advantage.

\subsubsection{Case 3.1}
\textbf{Question: How many countries in the world have scouts?}
\begin{tcbraster}[
    raster columns=1,
    enhanced,
    sharp corners
]

\begin{tcolorbox}[
    title=Search-R1-7B,
    colback=bad,
    colframe=red!60,
    fonttitle=\bfseries,
    title style={fill=red!20}
]
\texttt{<think>}I need to find out how many countries in the world have scouts. I'll search for it.\texttt{</think>}

\hl{\texttt{<search>} how many countries in the world have scouts \texttt{</search>}}

\texttt{<information>}

Doc 1: (Title: List of World Organization of the Scout Movement members) The World Organization of the Scout Movement is the largest international scout organization and was established in 1922. It has 176 members.

Doc 2: (Title: Scouting Countries) This is a complete list of the countries where Scouting exists. The asterisks mark countries that have third-party operating agreements with the US.

Doc 3: (Title: Scouting Around The World: Facts, Countries Involved, And History) Scouting exists in over 216 countries and takes on many different forms! Each Scouting organization has its own unique culture.

Doc 4: (Title: World Scouting) The Asia-Pacific Scout Region fosters the Scout Movement in 24 countries. Of the over 57 million members worldwide, more than 35 million young people and over ...

Doc 5: (Title: Scouting in Other Countries 2025 - World Population Review) Just about every country in the world, with a few notable exceptions, has a scouting program. Although there are several scouting programs globally, only about ...

\texttt{</information>}

\texttt{<think>}I found out that the World Organization of the Scout Movement has 176 members. Now I need to find out how many countries in the world have scouts.\texttt{</think>}

\hl{\texttt{<search>} how many countries have scouts in the world \texttt{</search>}} \textcolor{red!70!black}{(Duplicated search !)}

\texttt{<information>}

Doc 1: (Title: List of World Organization of the Scout Movement members) The World Organization of the Scout Movement is the largest international scout organization and was established in 1922. It has 176 members.

Doc 2: (Title: Scouting Countries) This is a complete list of the countries where Scouting exists. The asterisks mark countries that have third-party operating agreements with the US.

Doc 3: (Title: Scouting Around The World: Facts, Countries Involved, And History) Scouting exists in over 216 countries and takes on many different forms! Each Scouting organization has its own unique culture.

Doc 4: (Title: World Scouting) The Asia-Pacific Scout Region fosters the Scout Movement in 24 countries. Of the over 57 million members worldwide, more than 35 million young people and over ...

Doc 5: (Title: Scouting in Other Countries 2025 - World Population Review) Just about every country in the world, with a few notable exceptions, has a scouting program. Although there are several scouting programs globally, only about ...

\texttt{</information>}

\texttt{<think>}I found out that scouting exists in over 216 countries. Now I can provide the answer.\texttt{</think>}

\texttt{<answer>} 216 \texttt{</answer>} \correct{}
\end{tcolorbox}

\begin{tcolorbox}[
    title=TAPO-7B,
    colback=good,
    colframe=blue!60,
    fonttitle=\bfseries,
    title style={fill=blue!20}
]
\texttt{<think>}
To answer this question, I must gather the relevant information about how many countries in the world have Scouting.
\texttt{</think>}

\texttt{<search>}number of countries that have Scouting\texttt{</search>}

\texttt{<response>}

Doc 1: (Title: List of World Organization of the Scout Movement members) The World Organization of the Scout Movement is the largest international scout organization and was established in 1922. It has 176 members.

Doc 2: (Title: Scouting Countries) This is a complete list of the countries where Scouting exists. The asterisks mark countries that have third-party operating agreements with the US.

Doc 3: (Title: Scouts - World Scouting) There are currently more than 50 million Scouts, both young people and adults, across 216 countries and territories. Internationally recognised National Scout ...

Doc 4: (Title: Scouting in Other Countries 2025 - World Population Review) Just about every country in the world, with a few notable exceptions, has a scouting program. Although there are several scouting programs globally, only about ...

Doc 5: (Title: Scouting Around The World: Facts, Countries Involved, And History) Scouting exists in over 216 countries and takes on many different forms! Each Scouting organization has its own unique culture.

\texttt{</response>}

\texttt{<answer>}216\texttt{</answer>} \correct{}
\end{tcolorbox}

\end{tcbraster}

\subsubsection{Case 3.2}
\textbf{Question: Who has won the CMA entertainer of the year the most?}
\begin{tcbraster}[
    raster columns=1,
    enhanced,   
    sharp corners
]

\begin{tcolorbox}[
    title=Search-R1-7B,
    colback=bad,
    colframe=red!60,
    fonttitle=\bfseries,
    title style={fill=red!20}
]
\texttt{<think>}I need to find out who has won the CMA Entertainer of the Year the most. I'll search for it.\texttt{</think>}

\hl{\texttt{<search>} who has won the CMA Entertainer of the Year the most \texttt{</search>}}

\texttt{<information>}

Doc 1: (Title: Country Music Association Award for Entertainer of the Year) Garth Brooks holds the record for most wins in the category, with seven, while three-time recipient George Strait has a leading nineteen nominations. To ...

Doc 2: (Title: CMA Awards Winners for Entertainer of the Year Who Won Multiple ...) All the CMA Awards Winners for Entertainer of the Year Who Have Won Multiple Times · Barbara Mandrell, 2 · Hank Williams Jr., 2 · Vince Gill, 2.

Doc 3: (Title: These Artists Have The Most CMA Entertainer Of The Year Wins) Garth Brooks (seven wins) 91, 92, 97, 98, 2016, 2017, 2019. Garth Brooks. Garth has the most wins in the top category and even took himself ...

Doc 4: (Title: Every CMA Awards Entertainer of the Year Winner Ever - The Boot) Garth Brooks won twice in the early 1990s, twice again in the late '90s, twice again in 2016 and 2017, and then in 2019 -- seven times in all -- ...

Doc 5: (Title: Who Has Won the Most CMA Awards This Century? - Billboard) Brooks \& Dunn and Vince Gill are tied for the most CMA Awards wins with 18 each. But Brooks \& Dunn won most of their awards, and Gill won all of his, in the ...

\texttt{</information>}

\texttt{<think>}I found out that Garth Brooks holds the record for most wins in the CMA Entertainer of the Year category, with seven wins.\texttt{</think>}

\hl{\texttt{<search>} who has won the CMA Entertainer of the Year the most \texttt{</search>}} \textcolor{red!70!black}{(Duplicated search !)}

\texttt{<information>}

Doc 1: (Title: Country Music Association Award for Entertainer of the Year) Garth Brooks holds the record for most wins in the category, with seven, while three-time recipient George Strait has a leading nineteen nominations. To ...

Doc 2: (Title: CMA Awards Winners for Entertainer of the Year Who Won Multiple ...) All the CMA Awards Winners for Entertainer of the Year Who Have Won Multiple Times · Barbara Mandrell, 2 · Hank Williams Jr., 2 · Vince Gill, 2.

Doc 3: (Title: These Artists Have The Most CMA Entertainer Of The Year Wins) Garth Brooks (seven wins) 91, 92, 97, 98, 2016, 2017, 2019. Garth Brooks. Garth has the most wins in the top category and even took himself ...

Doc 4: (Title: Every CMA Awards Entertainer of the Year Winner Ever - The Boot) Garth Brooks won twice in the early 1990s, twice again in the late '90s, twice again in 2016 and 2017, and then in 2019 -- seven times in all -- ...

Doc 5: (Title: Who Has Won the Most CMA Awards This Century? - Billboard) Brooks \& Dunn and Vince Gill are tied for the most CMA Awards wins with 18 each. But Brooks \& Dunn won most of their awards, and Gill won all of his, in the ...

\texttt{</information>}

\texttt{<think>}I found out that Garth Brooks has won the CMA Entertainer of the Year the most, with seven wins.\texttt{</think>}

\texttt{<answer>} Garth Brooks \texttt{</answer>} \correct{}
\end{tcolorbox}

\begin{tcolorbox}[
    title=TAPO-7B,
    colback=good,
    colframe=blue!60,
    fonttitle=\bfseries,
    title style={fill=blue!20}
]
\texttt{<think>}
To answer this question, I need to gather relevant information about who has won the CMA Entertainer of the Year the most.
\texttt{</think>}

\texttt{<search>}who has won the CMA Entertainer of the Year the most\texttt{</search>}

\texttt{<response>}

Doc 1: (Title: Country Music Association Award for Entertainer of the Year) Garth Brooks holds the record for most wins in the category, with seven, while three-time recipient George Strait has a leading nineteen nominations. To ...

Doc 2: (Title: CMA Awards Winners for Entertainer of the Year Who Won Multiple ...) All the CMA Awards Winners for Entertainer of the Year Who Have Won Multiple Times · Barbara Mandrell, 2 · Hank Williams Jr., 2 · Vince Gill, 2.

Doc 3: (Title: These Artists Have The Most CMA Entertainer Of The Year Wins) Garth Brooks (seven wins) 91, 92, 97, 98, 2016, 2017, 2019. Garth Brooks. Garth has the most wins in the top category and even took himself ...

Doc 4: (Title: Who Has Won the Most CMA Awards This Century? - Billboard) Brooks \& Dunn and Vince Gill are tied for the most CMA Awards wins with 18 each. But Brooks \& Dunn won most of their awards, and Gill won all of his, in the ...

Doc 5: (Title: Every CMA Awards Entertainer of the Year Winner Ever - The Boot) Garth Brooks won twice in the early 1990s, twice again in the late '90s, twice again in 2016 and 2017, and then in 2019 -- seven times in all -- ...

\texttt{</response>}

\texttt{<answer>}Garth Brooks\texttt{</answer>} \correct{}
\end{tcolorbox}

\end{tcbraster}

\section{Limitations}
\label{app:limit}
\subsection{GPU Bubble During Training}
\begin{figure}[H]
    \centering
    \includegraphics[width=\linewidth]{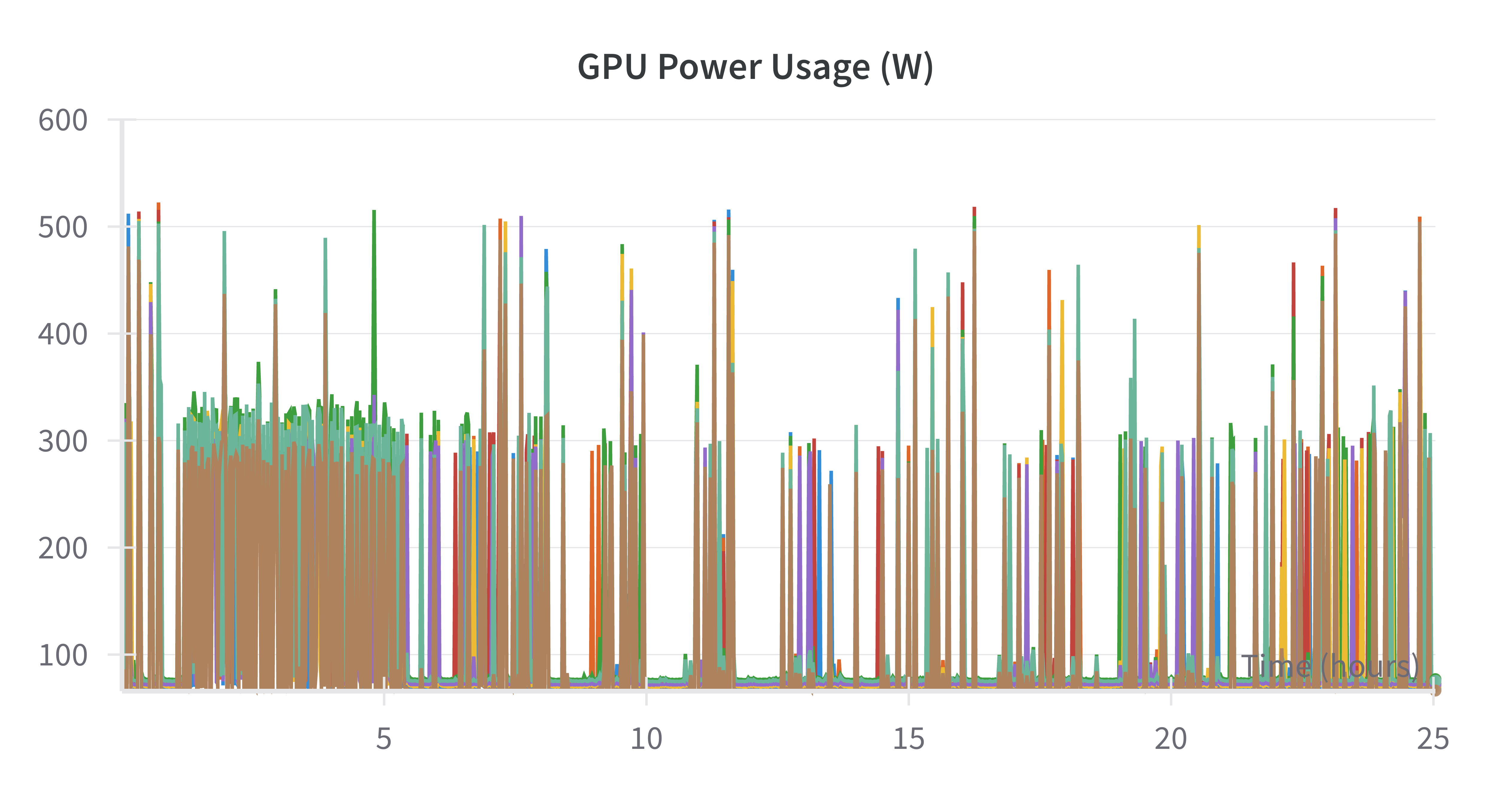}
    \includegraphics[width=\linewidth]{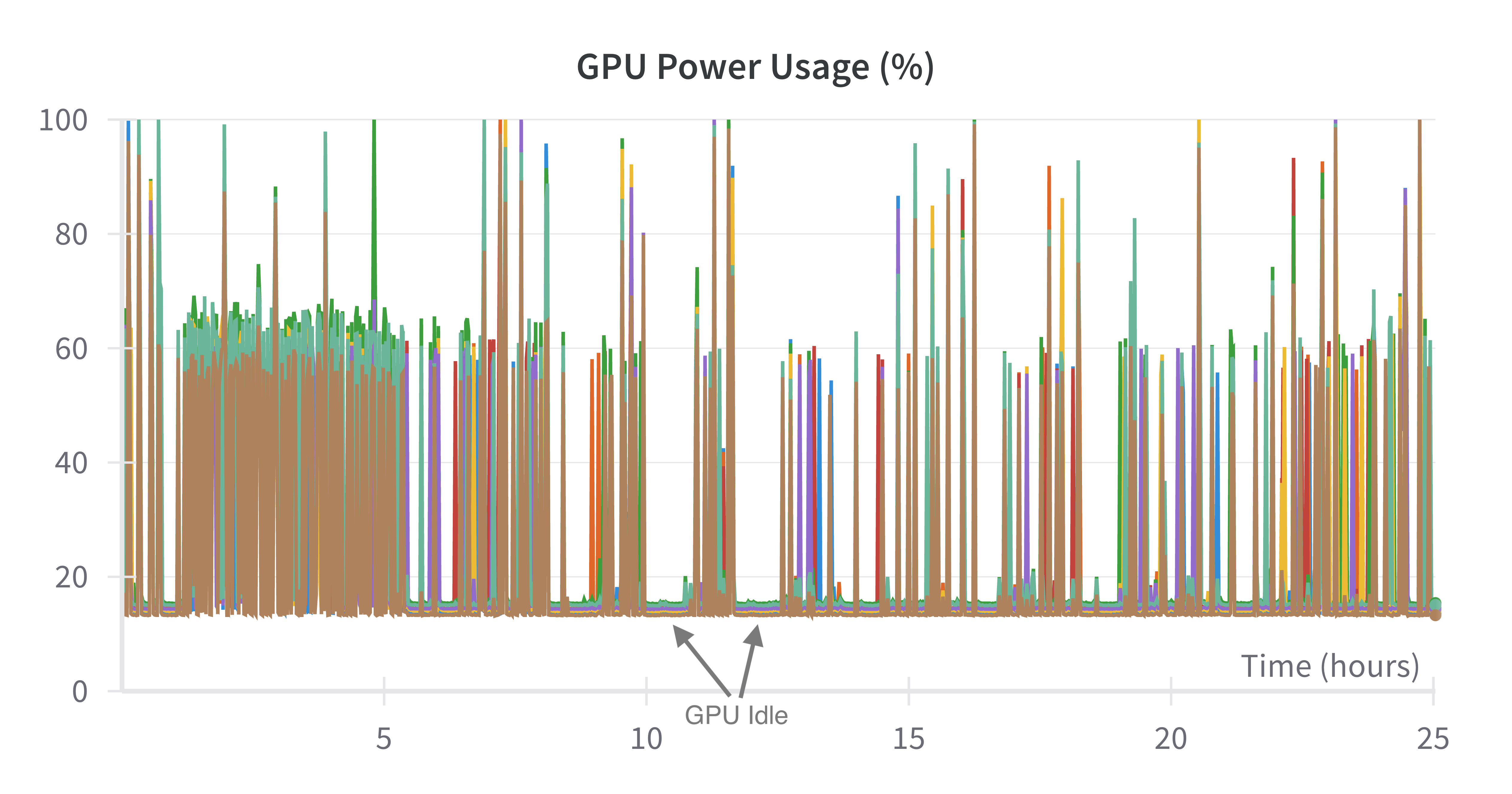}
    \caption{GPU power usage during training of TAPO in Qwen2.5-7B. Massive tool invocations cause severe GPU idleness.}
    \label{fig:gpu_power}
\end{figure}

The GPU bubble phenomenon poses a significant challenge in TAPO's training pipeline. Consider training a 7B parameter model with a batch size of 128: when each question generates 8 rollouts and assuming at least one tool invocation per rollout, the system must handle over 1,000 external requests per batch. Despite our optimizations to the tool invocation mechanism, this massive volume of requests introduces substantial network I/O overhead and waiting time during training. Consequently, these frequent synchronization points create prolonged periods of GPU idleness, significantly impacting the overall training efficiency (as shown in Figure \ref{fig:gpu_power}).

In future work, we plan to optimize the training pipeline by overlapping tool invocation I/O with language model inference, thereby hiding the latency through concurrent execution.

\subsection{Low Generalization Ability in Small Models}
While the 7B model demonstrates improved comprehensive capabilities, smaller models often exhibit poor performance on out-of-domain datasets. We attribute this limitation to the difficulty small models face in simultaneously mastering multiple tools, particularly code generation. For future work, we plan to implement a cold-start supervised fine-tuning (SFT) phase for small models prior to TAPO's reinforcement learning stage.

\end{appendices}

\end{document}